\documentclass{article}

\usepackage[final]{neurips_2025}

\usepackage[utf8]{inputenc} 
\usepackage[T1]{fontenc}    
\usepackage{hyperref}       
\usepackage{url}            
\usepackage{booktabs}       
\usepackage{amsfonts}       
\usepackage{nicefrac}       
\usepackage{microtype}      
\usepackage{xcolor}         
\usepackage{amsmath}
\usepackage{multirow}
\usepackage{color, colortbl}
\usepackage{soul}
\usepackage{booktabs} 
\usepackage{enumitem}
\usepackage{boldline}
\usepackage{array}
\usepackage{graphicx}
\usepackage[caption = false]{subfig}
\usepackage{pdfpages}
\usepackage{algorithm}
\usepackage{algorithmic}
\usepackage{diagbox}
\usepackage{url}
\usepackage{flushend}
\usepackage{xcolor}
\usepackage{hhline}
\usepackage{amsthm}
\usepackage{bbm}
\usepackage{caption}
\usepackage{amssymb}

\definecolor{LightCyan}{rgb}{0.88,1,1}
\definecolor{Gray_d}{gray}{0.8}
\definecolor{Gray_l}{gray}{0.9}

\DeclareMathOperator*{\argmin}{arg\,min}

\newcommand{\model}{DynaInfer}

\title{Environment Inference for Learning Generalizable Dynamical System}

\author{%
Shixuan Liu$^{1}$\quad Yue He$^{2}$\thanks{Corresponding authors} \quad Haotian Wang$^1$ \quad Wenjing Yang$^1$ \quad Yunfei Wang$^3$ \\
\textbf{Peng Cui}$^{4*}$ \quad \textbf{Zhong Liu}$^5$\\
$^1$College of Computer Science and Technology, National University of Defense Technology\\
$^2$School of Information, Renmin University of China\\
$^3$College of Systems Engineering, National University of Defense Technology\\
$^4$Department of Computer Science and Technology, Tsinghua University\\
$^5$Laboratory for Big Data and Decision, National University of Defense Technology\\
\texttt{szftandy@hotmail.com}, \texttt{hy865865@gmail.com}\\
\texttt{\{wanghaotian13,wenjing.yang,wangyunfei,liuzhong\}@nudt.edu.cn}\\
\texttt{cuip@tsinghua.edu.cn}
}

\newtheorem{theorem}{Theorem}[section]

\newtheorem{proposition}[theorem]{Proposition}
\theoremstyle{definition}

\theoremstyle{remark}
\newtheorem{remark}{Remark}[section]

\newcommand{\bfe}{\boldsymbol{e}}
\newcommand{\bfphi}{\boldsymbol{\phi}}

\newcommand{\calH}{\mathcal{H}}
\newcommand{\calA}{\mathcal{A}}

\begin{document}

\maketitle

\begin{abstract}
Data-driven methods offer efficient and robust solutions for analyzing complex dynamical systems but rely on the assumption of I.I.D. data, driving the development of generalization techniques for handling environmental differences. These techniques, however, are limited by their dependence on environment labels, which are often unavailable during training due to data acquisition challenges, privacy concerns, and environmental variability, particularly in large public datasets and privacy-sensitive domains.
In response, we propose \model, a novel method that infers environment specifications by analyzing prediction errors from fixed neural networks within each training round, enabling environment assignments directly from data. We prove our algorithm effectively solves the alternating optimization problem in unlabeled scenarios and validate it through extensive experiments across diverse dynamical systems. Results show that \model~outperforms existing environment assignment techniques, converges rapidly to true labels, and even achieves superior performance when environment labels are available.
\end{abstract}

\section{Introduction}
\label{sec:intro}

Data-driven approaches, especially neural networks, offer a powerful alternative or complement to traditional physics-based methods for understanding complex dynamical systems~\citep{brunton2016discovering}. Neural network-based emulators are particularly valuable for their ability to provide fast, cost-effective approximations of complex simulations~\citep{duraisamy2019turbulence,li2020fourier}, making them especially useful in scenarios where the underlying physics are poorly understood or misinterpreted, or where external disturbances are difficult to model~\citep{yin2021leads,sirignano2018dgm}. These emulators are adept at handling large sets of variables and solving problems that are challenging for conventional solvers. Recent advancements in deep learning, along with innovative methods for modeling temporal and spatio-temporal systems, have led to a significant increase in applications across various fields, ranging from simple Hamiltonian dynamics to more complex areas like fluid dynamics and climatology~\citep{shaier2021data,de2019deep}.


While recent advancements have shown promising results, they often rely on the assumption that abundant, static data are available to satisfy the independent and identically distributed (IID) hypothesis.
However, this assumption is frequently violated in practice due to challenges in data collection, associated costs, and environmental changes driven by exogenous factors~\citep{madec2017nemo,neic2017efficient}. 
Recent work in dynamical systems addresses this by introducing a multi-environment setting, where trajectories follow distinct dynamics across environments. These studies developed generalization methods that learn a shared global component while accounting for environment-specific variations, avoiding the limitations of underperforming averaged models~\citep{yin2021leads,kirchmeyer2022generalizing}.

\begin{figure}[H]
\centering
\includegraphics[width=0.82\linewidth, keepaspectratio]{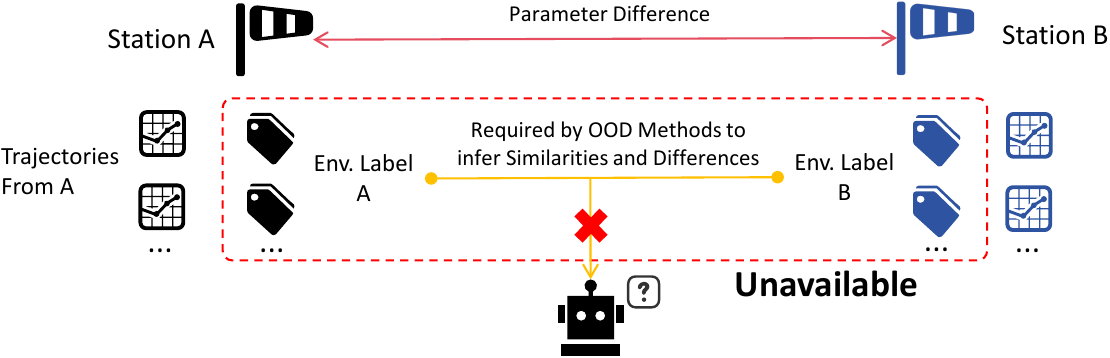}
\caption{Environment labels, required by current generalization methods, are often unavailable.}
\label{fig_model}
\end{figure}

Nevertheless, a key limitation of many generalization techniques is their reliance on partitioning datasets across distinct domains or environments, which are assumed to capture underlying variations. These environment labels enable algorithms to identify and exploit both similarities and differences across environments. However, obtaining such environment labels during training is often challenging due to data acquisition difficulties or privacy constraints. For example, in scientific research, data may be collected over time under uncontrolled or unknown conditions~\citep{willard2020integrating}. In ecological studies, critical environmental parameters such as temperature or rainfall may vary unpredictably or remain unrecorded.~\citep{ben2009robust}. Similarly, when aggregating data from multiple sources, environment labels are frequently lost or omitted, a common issue in large public datasets~\citep{srivastava2020robustness}. Furthermore, in privacy-sensitive domains like healthcare, finance, or social networking, access to environment-specific information is often restricted~\citep{lahoti2020fairness}. These limitations highlight the need for generalization methods that do not depend on explicit environment labels.


To address the challenge of unknown environment labels, we propose a novel approach that infers environment specifications by leveraging the key insight that trajectories within the same environment share consistent dynamics and exhibit similar prediction losses under the same neural network. This inherent consistency enables us to automatically derive meaningful environment assignments directly from the training data. We introduce an environment inference objective designed for dynamical systems, which minimizes environment-specific prediction losses.
Using fixed neural networks, we first infer environments and then iteratively refine these networks with the inferred environments, ultimately learning a generalizable dynamical system.



Our model identifies environment labels directly from mixed trajectories of dynamical systems, facilitating the training of off-the-shelf generalization algorithms in scenarios where such labels are absent. Importantly, our findings demonstrate that inferring environments from mixed sequence data can improve the performance of generalization strategies, even compared to cases where environments are manually assigned.

Our main contributions are as follows:
\begin{itemize}
    \item We present the first investigation into the challenge of unlabeled environment conditions in the context of learning generalizable dynamical systems, and propose a general framework named \model~that utilizes the prediction loss to accurately infer latent environment labels from mixed sequence data\footnote{Code is available at \url{https://github.com/shixuanliu-andy/DynaInfer}}.


    \item We theoretically establish that our algorithm effectively solves the alternating optimization problem without requiring environment labels, demonstrating its capacity to discern heterogeneous environments and infer generalizable mechanisms.
    
    \item We examine the efficacy of \model~through experiments in both in-domain settings and adaptation scenarios using three representative dynamical systems. Results confirm that the environment labels assigned by \model~converge rapidly to the true labels.

\end{itemize}

The remainder of this paper is structured as follows. Section 2 clarifies the problem definition. Section 3 introduces our framework and provides the theoretical underpinnings. Section 4 details the experimental setup and discusses the results. Related work is reviewed in Section 5, and Section 6 concludes the paper.
\section{Problem Definition}
\subsection{Dynamical Systems}
We examine dynamical systems determined by unidentified differential equations evolving over time, expressed as,
\begin{equation}
    \frac{dx_t}{dt}= f(x_t)
\end{equation}
where $t \in \mathbb{R}$ is the time index within a time interval $I=[0,T]$, and $x_t$ is a time-variant state within a bounded set $\mathcal{A}$. 
The evolution function $f: \mathcal{A} \rightarrow T\mathcal{A}$ maps $x_t$ to its temporal derivative in the tangent space $T\mathcal{A}$ and belongs to a class of vector fields $\mathcal{F}$.

In this paper, we consider both ordinary differential equation (ODE) and partial differential equation (PDE).
For ODEs, $\mathcal{A} \subset \mathbb{R}^d$; for PDEs, $\mathcal{A}$ represents a $d'$-dimensional vector field within a bounded spatial domain (such as 2D or 3D Euclidean space) denoted as $S \subset \mathbb{R}^{d'}$. 
The function $f$ characterizes the data distribution of trajectories $\mathcal{T}$. 
Trajectories initiated from $x_0 \sim p(X_0)$ are computed by integrating the derivatives: $x_t = x_0 + \int_0^t f(x_u)du, \forall t \in I$.

\subsection{Multi-Environment Dynamical Systems Learning}

In contrast to the standard expected risk minimization (ERM) framework, which assumes i.i.d. trajectories, the multi-environment learning problem involves learning trajectories from $M$ different environments.
In each environment $e \in [M] = \{1, 2, \dots, M\}$, the trajectories are governed by unique differential equations described by function $f_e$.
Specifically, consider $N$ trajectories $\{x^1, x^2, \dots, x^N\}$, where each trajectory $x^i$ is associated with an environment $e_i \in [M]$. The dynamics of each trajectory $x^i$ are thus modeled by the differential equation $d x^i_t / dt = f_{e_i}(x^i_t)$.
The set of environments for all trajectories is denoted by $\bfe = \{e_1, e_2, \dots, e_N\} \in [M]^N$.



In multi-environment learning, the goal is to enhance traditional ERM methods by exploiting both the commonalities and disparities across diverse environments. To this end, the dynamics is decomposed into two components: a global component shared across all environments, parameterized by $\theta$, and an environment-specific component, parameterized by $\phi_e$ for each environment $e$. The set of environment-specific parameters is denoted by $\bfphi = \{\phi_e\}_{e \in [M]}$. Consequently, the dynamics of each trajectory $x^i$ are parameterized by both the universal and environment-specific parameters,

$$
\frac{d x^{i}_t}{d t} = h\left(x^{i}_t; \theta, \phi_{e}\right).
$$

This parametrization entails a decomposition that can be implemented either functionally or parametrically.
The functional decomposition, expressed as $h\left(x^{i}_t; \theta, \phi_{e}\right)=f_{\theta}(x^{i}_t) + g_{\phi_{e}}(x^{i}_t)$, distinguishes between a shared function $f_{\theta}$ and an environment-specific function $g_{\phi_{e}}$~\citep{yin2021leads}. Alternatively, the parametric decomposition integrates the environment-specific parameters directly, formulated as $h\left(x^{i}_t; \theta, \phi_{e}\right)=f_{\theta + \phi_{e}}(x^{i}_t)$~\citep{kirchmeyer2022generalizing}.
Intuitively, the key ingredient for multi-environment learning is that $\theta$ should encapsulate the maximal shared dynamics, whereas $\phi_{e}$ should exclusively reflect the unique characteristics of each environment $e$ not described by $\theta$. However, directly optimizing both parameters poses an ill-posed problem, often resulting in trivial solutions where the global component learns nothing meaningful.
To counteract this, the regularization term $\Omega(\phi_{e})$ is introduced to effectively penalize $\phi_e$, thereby facilitating learning in the global component.
Consequently, with the information about the environments $\bfe = \{e_1, e_2, \dots, e_N\}$, the loss function is given by,

\begin{equation} \label{eq:loss}
R_{\bfe}(\theta, \bfphi) = \sum_{i=1}^N \int_{t\in I} \left\|\frac{dx^{i}_t}{dt} - h\left(x^{i}_t; \theta, \phi_{e_i}\right)\right\|_2^2dt + \lambda \sum_{e=1}^M \Omega(\phi_{e}).
\end{equation}

The first term evaluates the regression precision of the parameterized function $h(\cdot; \theta, \phi_e)$. The ground truth vector field (VF) is not explicitly known and derived from trajectory data. Using the learned VF, a simulated trajectory is generated and used to calculate the regression loss by referring to real trajectories during training. 
The term $\Omega(\phi_e)$ serves as a regularization term for $\phi_e$, with $\lambda$ controlling the intensity of the regularization.

\subsection{Environment Inference for Multi-Environment Learning}
In many real-world scenarios, the environment label for a trajectory sample is unknown. We aim to infer an environment assignment for each sample that maximizes the model's generalization ability across different environments. 
To achieve this goal, we reformulate the learning objective into an optimization problem contingent on a specific environment assignment $\bfe$. 
Specifically, our aim is to learn the environment assignment $\hat{\bfe} = \{\hat{e}^1, \hat{e}^2, \dots, \hat{e}^N\} \in [M]^N$ for each trajectory to effectively optimize Equation (\ref{eq:loss}). The overall objective is defined as follows:
\begin{equation} 
\label{eq:loss-argmin}
\hat{\bfe}^*, \theta^*, \bfphi^* = \argmin_{\hat{\bfe}, \theta, \bfphi} R_{\hat{\bfe}}(\theta, \bfphi).
\end{equation}



In this paper, we explore a particularly challenging scenario where the total number of training environments $M$ is also unknown. 
We investigate the development of a practical model that maintains favourable performance even when the exact number of true environments is unknown.
\section{The \model~Framework}
In this section, we introduce our framework that operates on field functions without prior domain knowledge, proving especially effective in dynamical systems where exogenous factors are unobserved and in situations where relevant environmental information is unclear or absent. 
While some clustering methods infer labels for CV data, they operate on finite-dimensional vectors in Euclidean space, which drastically differs from field functions, making them inapplicable.


The optimization challenge in Equation~\eqref{eq:loss-argmin} is primarily due to the inherently discrete nature of the environment assignments $\hat{\bfe}$, which take values in the set $[M]$. This discrete categorization impedes the direct application of traditional gradient descent methods, which are typically designed for continuous parameter spaces. 
To effectively address this challenge, we develop a dual iterative strategy that concurrently updates the environment assignments $\hat{\bfe}$ and the model parameters $\theta$, $\bfphi$. 
The first step in our approach centers on inferring environment labels by analyzing the prediction errors of the trajectories output by the neural network during the current training round.
This analysis serves as a diagnostic tool to uncover critical discrepancies that signify distinct dynamical environments.
Following this, the second step entails refining the neural network parameters based on the newly inferred environment assignments in an unbiased manner, enabling the neural network to precisely adapt to the unique characteristics of each identified environment.
Through this adaptive refinement, our model progressively enhances its accuracy and generalization capability across different dynamic settings.
The complete method is detailed in Algorithm~\ref{alg:example} and is visually depicted in Figure~\ref{fig_model}.

\noindent\begin{minipage}{.62\linewidth} 
\begin{algorithm}[H]
\caption{\model~framework}
\label{myalg}
\label{alg:example}
\begin{algorithmic}[1]
\STATE {\bfseries Input:} Randomly initialized $\theta, \bfphi = \{\phi_e\}_{e \in [M]}$, assumed number of environments $M$, total rounds $T_r$
\STATE $\theta^{(0)}, \bfphi^{(0)} \leftarrow \theta, \bfphi$
\FOR{$r \leftarrow 1$ to $T_r$}
    \STATE Update $\hat{\bfe}^{(r)}$ based on Equation~\eqref{eq:e-updating}
    \STATE Update $\theta^{(r)}, \bfphi^{(r)}$ based on Equation~\eqref{eq:theta-phi-updating}
\ENDFOR
\STATE {\bfseries Output:} $\theta, \bfphi$.
\end{algorithmic}
\end{algorithm}
\end{minipage}
\noindent \begin{minipage}{.36\linewidth}
\begin{figure}[H]
\centering
\includegraphics[width=\textwidth, keepaspectratio]{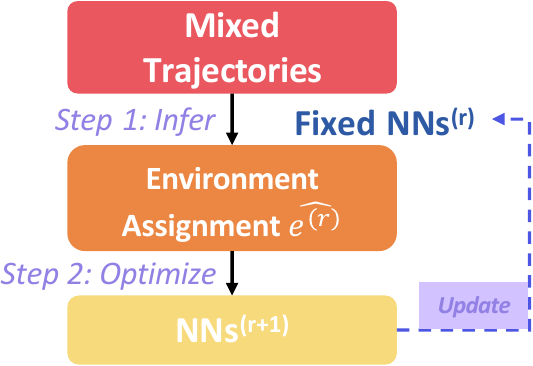}
\caption{Model Framework.}
\label{fig_model}
\end{figure}
\end{minipage}

\subsection{Bias-aware Environment Assignment}

The environment inference step receives a single dataset as input and generates a partition of the data into multiple environments. Intuitively, trajectories originating from the same environment adhere to consistent dynamics. Employing the same neural network to model these trajectories should yield similar estimation error across them, reflecting a coherence in their dynamic parameters.



Upon examination, we observe that the optimization framework defined in Equation (\ref{eq:loss}) shares conceptual similarities with classical centroid-based clustering methodologies, although the latter generally operate in Euclidean space.
In K-means clustering, the primary goal is to minimize the within-cluster sum of squares, often referred to as cluster inertia~\citep{sun2012regularized}. 
This minimization effort concentrates on reducing the distances between the points within each cluster and their corresponding centroid, which typically converges to a local optimum. 
This characteristic enables K-means to efficiently delineate distinct and compact clusters, capturing the core essence of data distribution with respect to spatial proximity.


This insight prompts us to explore a conceptual analogy wherein the neural network that minimizes the loss most effectively operates analogously to a "centroid" for a cluster of trajectories within the same dynamic environment. We characterise the distance between a trajectory (data point) and the network (centroid) by the regression loss of the trajectory using the network. 
Initially, with a randomly initialized network—analogous to a randomly initialized centroid in K-means clustering—we assign each trajectory a label based on the minimal prediction loss calculated from all available networks. Subsequently, we refine this "centroid" by optimizing it to minimize the loss as specified in Equation (\ref{eq:loss}). Through this iterative optimization process, we can achieve the objective stated in Equation (\ref{eq:loss-argmin}).



More specifically, at round $r$, given the fixed network parameters from the previous iteration $\theta^{(r - 1)}, \bfphi^{(r - 1)}$, the environment assignment $\hat{e}_i^{(r)}$ is updated through the following process, 
\begin{equation} 
\label{eq:e-updating}
\hat{e}_i^{(r)} = \argmin_{e \in [M]} \int_{t\in I} \left\|\frac{dx^{i}_t}{dt} - h\left(x^{i}_t; \theta^{(r - 1)}, \phi_{e}^{(r - 1)}\right)\right\|_2^2dt.
\end{equation}
If multiple solutions exist for Equation~\eqref{eq:e-updating} and $\hat{e}_i^{(r-1)}$ minimizes it, we retain this assignment for the next round, i.e., $\hat{e}_i^{(r)} = \hat{e}_i^{(r - 1)}$. This approach ensures the validity of a constant loss reduction (to be stated in Proposition~\ref{prop:convergence}).

\subsection{Assignment-driven Optimization}

After assigning trajectories to specific clusters, we proceed to update the conceptual centroid by optimizing network parameters. In the K-means algorithm, centroids are recalculated by averaging the positions of all points within each cluster. Similarly, our method updates network parameters by considering the mean estimation error over trajectories within a cluster, ensuring unbiased contributions from each trajectory. This approach not only improves the representational accuracy of each cluster but also enables the network to dynamically adapt to the underlying structure of the trajectories, thereby enhancing the efficacy and reliability of our learning process in unlabeled scenarios.
Therefore, the parameters $\theta^{(r)}$ and $\bfphi^{(r)}$ are given by:
\begin{equation} \label{eq:theta-phi-updating}
    \theta^{(r)}, \bfphi^{(r)} = \argmin_{\theta, \bfphi} R_{\hat{\bfe}^{(r)}}(\theta, \bfphi).
\end{equation}

\subsection{Theoretical Property}

We begin by demonstrating that Algorithm~\ref{alg:example} effectively optimizes Equation~\eqref{eq:loss-argmin}.
\begin{proposition} 
\label{prop:convergence}
    For all rounds $1 \le r < T_r$, we must have
    $$
    R_{\hat{\bfe}^{(r + 1)}}\left(\theta^{(r + 1)}, \bfphi^{(r + 1)}\right) \le R_{\hat{\bfe}^{(r)}}\left( \theta^{(r)}, \bfphi^{(r)}\right).
    $$
    Furthermore, suppose the space of $\argmin_{\theta, \bfphi}R_{\hat{\bfe}}(\theta, \bfphi)$ is finite for all $\hat{\bfe} \in [M]^N$. Then there exists a constant $C>0$ such that if $r > 1$ and $R_{\hat{\bfe}^{(r + 1)}}(\theta^{(r + 1)}, \bfphi^{(r + 1)}) < R_{\hat{\bfe}^{(r)}}( \theta^{(r)}, \bfphi^{(r)})$, we must have
    $$
    R_{\hat{\bfe}^{(r + 1)}}\left(\theta^{(r + 1)}, \bfphi^{(r + 1)}\right) \le R_{\hat{\bfe}^{(r)}}\left( \theta^{(r)}, \bfphi^{(r)}\right) - C.
    $$
\end{proposition}

\begin{remark}
Given the assumptions made in prior works~\citep{yin2021leads,kirchmeyer2022generalizing} that $h(\cdot; \theta, \phi_e)$ is linear with respect to $\theta$ and $\phi_e$, and that $\Omega(\phi_e)$ is strictly convex with respect to $\phi_e$, it follows logically that the space of $\argmin_{\theta, \bfphi} R_{\hat{\bfe}}(\theta, \bfphi)$ is finite for all $\hat{\bfe} \in [M]^N$, as is evident from Equation (\ref{eq:loss}).
The proof is provided in Appendix~\ref{sec:proof-convergence}.
\end{remark}

This proposition demonstrates that, as long as the loss in consecutive rounds of Algorithm~\ref{alg:example} decreases, the loss must decrease by a constant $C > 0$.

\section{Experiments}

Our experiments investigate three dynamical systems governed by specific differential equations: an ODE for biological modeling, PDEs for reaction-diffusion in chemistry, and the Navier-Stokes equations for incompressible fluid dynamics. These complex, nonlinear systems test our method's ability to classify spatio-temporal patterns and physical laws across diverse environments.

\subsection{Environment Specification}
We provide a basic introduction to the datasets here, with detailed descriptions in Appendix~\ref{appendix_environments}.
\noindent \textbf{Lotka-Volterra (LV)~\citep{lotka1925elements}} The system models the dynamics between a prey-predator pair in an ecosystem, captured by the following ODE:
\begin{equation*}
dm/dt = \alpha m - \beta mn, dn/dt = \delta mn - \gamma n
\end{equation*}
where $m$ and $n$ represent the population densities of the prey and predator, respectively, and $\alpha$, $\beta$, $\delta$, and $\gamma$ are the interaction parameters between the two species.

\noindent \textbf{Gray-Scott (GS)~\citep{pearson1993complex}} The model uses simple reaction-diffusion equations to effectively study complex pattern formation in chemical and biological systems, following underlying PDE dynamics:
\begin{equation*}
\begin{aligned}
& \partial m/ \partial t = D_m\Delta m - mn^2 + F(1-m), \\
& \partial n/ \partial t = D_n\Delta n - mn^2 - (F+k)n.    
\end{aligned}
\end{equation*}
where $m$ and $n$ represent the concentrations of two chemical components in the spatial domain $S$ with periodic boundary conditions; $D_m$ and $D_n$ are their constant diffusion coefficients; and $F$ and $k$ are the reaction parameters that govern the spatio-temporal dynamic patterns.

\noindent \textbf{Navier-Stokes (NS)~\citep{li2020fourier}} The Navier-Stokes PDE describes the motion of viscous fluid substances:

\begin{equation*}
\partial m/ \partial t = -n \nabla m + \nu \Delta m + \xi, \nabla v = 0
\end{equation*}
where $n$ is the velocity field, $m = \nabla \times n$ is the vorticity, both $n$ and $m$ lie in a spatial domain $S$ with periodic boundary conditions, $\nu$ is the viscosity (fixed at $1e^{-3}$), and $\xi$ is the constant forcing term in the domain $S$.


\subsection{Experimental Setting and Baselines}
\noindent \textbf{Settings.}  
We evaluate \model~in two distinct settings: in-domain generalization on $\mathcal{E}_o$ and adaptation to new environments in $\mathcal{E}_u$, with $\mathcal{E}_o$ and $\mathcal{E}_u$ hosting disjoint environments. 
For in-domain experiments, both training and testing occur on $\mathcal{E}_o$. 
At test time, environment labels are also not provided and are instead inferred from the prediction bias over an initial segment of the trajectory (less than $2\Delta t$ for practical reasons).
For adaptation experiments, we follow standard domain adaptation practice: initial training on the source domain $\mathcal{E}_o$ is followed by fine-tuning and testing on the target domain $\mathcal{E}_u$, where environment labels are provided.

\noindent \textbf{Dataset Preparation.}
For in-domain experiments, we generate four LV trajectories in each of nine environments, ten GS trajectories in each of three environments, and eight NS trajectories in each of four environments.
For adaptation experiments, we simulate the same number of trajectories per environment, conducting finetuning in two additional environments $e \in \mathcal{E}_u$. All dynamic environment parameters are detailed in Appendix~\ref{appendix_environments}.
For evaluation, we sample 32 trajectories per environment, initialized according to the underlying distribution $p(x_0)$. The LV and GS data are generated using the DOPRI5 solver~\citep{dormand1980family,harris2020array}, while the NS data is simulated with the pseudo-spectral method as in~\citep{li2020fourier}.

\noindent \textbf{Baselines.} 
We explore three potential strategies for assigning environment labels in the absence of environmental information, compared to our method (\model): grouping all samples into a single environment (All in One), assigning a distinct environment label to each sample (One per Env), and random assignment (Random).
Additionally, we consider an "Oracle" assignment method where labels are fully known during training, bringing the total to five labeling strategies.
Furthermore, we consider three base models for dynamical system generalization: LEADS~\citep{yin2021leads}, CoDA-$l_1$, and CoDA-$l_2$~\citep{kirchmeyer2022generalizing}. We utilize the neural network architectures and parameter configurations as described in their papers for each type of dynamic system.
By combining these assignment methods with base models, we generate fifteen distinct methods for evaluation.
In adaptation experiments, during fine-tuning, LEADS and CoDA adhere to the protocol described in their papers, by fixing the shared components or parameters and rendering only the $\mathcal{E}_u$-specific components trainable. All neural network architectures, optimizers, and parameters for the base models are configured as described in their respective papers.

\noindent \textbf{Metrics.} To rigorously evaluate predictive accuracy in dynamical system learning, we adopt two complementary metrics: Mean Squared Error (MSE) and Mean Absolute Percentage Error (MAPE), averaged over 5 independent runs. 

\subsection{Experimental Results}
\subsubsection{In-domain Generalization Results} 
The in-domain generalization results detailed in Table~\ref{table_main} illustrate the performance implications of various assignment strategies. 
We observe that the "All in One" and "Random" assignment strategies consistently underperform across multiple datasets and baseline models. 
While the "One per Env" strategy yields only mediocre results, it provides a viable initial approach in scenarios where no labels are available. Across all datasets, \model~significantly outperforms other assignment strategies. Furthermore, \model~consistently shows effectiveness across all tested base models and datasets, underscoring its robustness against diverse methods and datasets. 
Notably, \model~either matches or exceeds Oracle performance, particularly in complex PDE environments like GS and NS, suggesting that its bias-aware approach effectively compensates for not having access to the true labels available to Oracle.

\begin{figure*}[htb]
\centering
\includegraphics[width=\linewidth]{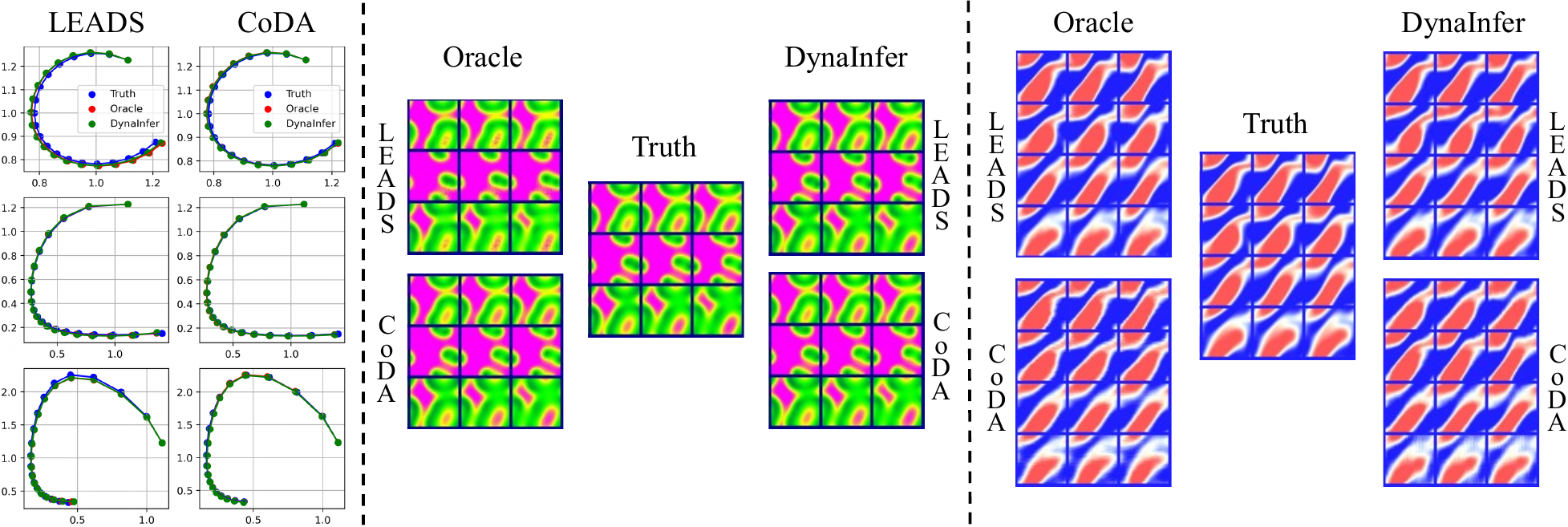}
\caption{Left: Predicted test trajectories from 3 environments vs. ground truth and Oracle for LV with 2 base models.  
Middle and Right: Predicted last 3 states for GS and NS, respectively, vs. ground truth and Oracle using 2 base models. Each environment is shown by row. For CoDA, we use the best Oracle result's CoDA version to save space. See Appendix~\ref{appendix_plot_dynamic} for all the visualizations.
}
\label{fig_dyn_all}
\end{figure*}

In Figure~\ref{fig_dyn_all}, DynaInfer's predicted states qualitatively align closely with the ground truth and Oracle, occasionally outperforming Oracle (e.g., in GS dataset with LEADS base model, where Oracle shows some jitters).

\begin{table*}[htb]
\centering
\renewcommand\arraystretch{1.2}
\resizebox{\linewidth}{!}{
\begin{tabular}{c|c|ccc|ccc|ccc}
\toprule[1.5pt]
\multirow{3}{*}{Data}  & \multirow{3}{*}{Assignment} & \multicolumn{3}{c|}{LEADS} & \multicolumn{3}{c|}{CoDA-$l_1$} & \multicolumn{3}{c}{CoDA-$l_2$} \\ \cline{3-11}

& & Train     & \multicolumn{2}{c|}{Test}  & Train & \multicolumn{2}{c|}{Test}   & Train       & \multicolumn{2}{c}{Test}    \\ 
& & MSE& MSE & MAPE & MSE & MSE & MAPE & MSE & MSE & MAPE
\\
\hline \multirow{5}{*}{LV} & All in One              & 7.17 E-2  & 7.41±0.02 E-2 & 49.22±1.84 & 7.14 E-2    & 7.40±0.01 E-2   &49.44±3.15   & 7.17 E-2    & 7.41±0.00 E-2  & 39.26±22.13  \\
& One per   Env           & 4.15 E-4  & 4.91±3.50 E-4 & 6.68±2.44 & 8.68 E-4    & 9.14±0.41 E-4  &5.67±1.01  & 8.18 E-4    & 8.43±0.39 E-4 &5.73±1.19   \\
& Random                  & 7.20 E-2  & 7.38±0.02 E-2 & 50.01±1.05 & 7.12 E-2    & 7.39±0.01 E-2    &48.87±1.81 & 7.09 E-2    & 7.39±0.00 E-2  &48.86±2.54  \\
& \cellcolor{Gray_l} \model  & \cellcolor{Gray_l} \textbf{4.74 E-5}  & \cellcolor{Gray_l} \textbf{7.93±2.49 E-5} & \cellcolor{Gray_l} \textbf{2.83±1.62} & \cellcolor{Gray_l} \textbf{9.57 E-5}    & \cellcolor{Gray_l} \textbf{1.83±3.40 E-4}  & \cellcolor{Gray_l} \textbf{3.27±2.36}  & \cellcolor{Gray_l} \textbf{1.71 E-4}    & \cellcolor{Gray_l} \textbf{1.82±3.07 E-4} & \cellcolor{Gray_l} \textbf{2.02±1.66}   \\
& \cellcolor{Gray_d} Oracle                  & \cellcolor{Gray_d} 4.55 E-5  & \cellcolor{Gray_d} 7.02±0.76 E-5 & \cellcolor{Gray_d} 1.78±0.10 &\cellcolor{Gray_d} 1.78 E-5    &\cellcolor{Gray_d} 3.19±0.24 E-5    &\cellcolor{Gray_d} 1.26±0.06 &\cellcolor{Gray_d} 1.99 E-5    &\cellcolor{Gray_d} 2.72±0.18 E-5 &\cellcolor{Gray_d} 1.21±0.08 \\ 

\hline \multirow{5}{*}{GS} & All in One              & 8.73 E-3 & 9.60±0.02 E-3& 3008.80±892.20 & 9.24 E-3&9.61±0.03 E-3& 4115.80±223.54&9.25 E-3&9.60±0.00 E-3 &3723.00±713.85  \\
& One per Env  & 1.38 E-3&1.65±0.54 E-3&173.44±59.16 &1.56 E-3 & 1.91±0.06 E-3&185.23±61.43 & 1.52 E-3& 1.87±0.02 E-3 &174.08±57.82  \\
& Random   & 8.78 E-3&9.36±0.20 E-3& 1403.50±119.50 &9.25 E-3&9.59±0.03 E-3 & 3958.25±682.38& 8.77 E-3&9.35±0.02 E-3 &3919.88±157.54 \\
& \cellcolor{Gray_l} \model   & \cellcolor{Gray_l} \textbf{3.60 E-5}&\cellcolor{Gray_l} \textbf{4.14±0.21 E-5} &\cellcolor{Gray_l} \textbf{117.57±33.90} &\cellcolor{Gray_l} \textbf{9.22 E-5} &\cellcolor{Gray_l} \textbf{1.23±0.41 E-4} & \cellcolor{Gray_l} \textbf{122.93±22.05}
 &\cellcolor{Gray_l} \textbf{6.69 E-5} &\cellcolor{Gray_l} \textbf{7.25±2.11 E-5}
 &\cellcolor{Gray_l} \textbf{112.52±14.15}\\
& \cellcolor{Gray_d} Oracle &\cellcolor{Gray_d} 7.73 E-5&\cellcolor{Gray_d}1.34±0.76 E-4
&\cellcolor{Gray_d}97.77±12.09 &\cellcolor{Gray_d}6.04 E-5&\cellcolor{Gray_d}9.60±3.91 E-5& \cellcolor{Gray_d}163.38±47.89 &\cellcolor{Gray_d}4.69 E-5&\cellcolor{Gray_d}7.04±1.84 E-5 &\cellcolor{Gray_d}138.86±16.55\\ 

\hline \multirow{5}{*}{NS} & All in One        & 5.34E-02&6.71±0.11 E-2
&239.70±14.78 &5.79E-02&6.64±0.11 E-2& 251.38±8.54 
 &6.17E-02&6.64±0.03 E-2 &255.26±9.11  \\
& One per Env  & 2.24E-02&4.11±0.14 E-2&169.48±9.68&3.45E-02&3.88±0.22 E-2& 161.04±10.73&2.31E-02&4.04±0.22 E-2
& 158.26±9.35
\\
& Random   & 3.06E-02&6.58±0.05 E-2&233.80 ± 8.44 &5.04E-02&6.58±0.05 E-2&247.95±4.59&5.78E-02&6.66±0.04 E-2 &254.47±6.40 \\
& \cellcolor{Gray_l} \model   & \cellcolor{Gray_l} \textbf{6.10E-04} &\cellcolor{Gray_l} \textbf{7.05±0.34 E-3} 
&\cellcolor{Gray_l} \textbf{77.29±10.18}
&\cellcolor{Gray_l} \textbf{1.23E-02} &\cellcolor{Gray_l} \textbf{1.62±0.18 E-2} 
&\cellcolor{Gray_l} \textbf{108.17±10.30 } &\cellcolor{Gray_l} \textbf{8.92E-04} &\cellcolor{Gray_l} \textbf{1.19±0.12 E-2}
&\cellcolor{Gray_l} \textbf{96.57±12.75}
\\
& \cellcolor{Gray_d} Oracle                  &\cellcolor{Gray_d} 2.59E-04&\cellcolor{Gray_d}6.55±1.34 E-3
&\cellcolor{Gray_d} 67.58±9.37
&\cellcolor{Gray_d}1.36E-02&\cellcolor{Gray_d}1.73±0.29 E-2
&\cellcolor{Gray_d}124.22±12.35 
&\cellcolor{Gray_d}7.11E-04&\cellcolor{Gray_d}9.46±0.51 E-3
&\cellcolor{Gray_d}91.06±5.85 
\\ 
\bottomrule[1.5pt]
\end{tabular}
}
\caption{In-domain Experiment Results on the LV, GS, and NS environments. Our approach consistently outperforms all non-oracle assignment methods, and beats oracle at times, demonstrating its effectiveness in modeling heterogeneous environments and generalizing across dynamical systems.
}
\label{table_main}
\end{table*}

\subsubsection{Domain Adaptation Results} 

The results in Table~\ref{table_adapt} demonstrate the performance of different assignment strategies under the domain adaptation setting. The "One per Env" strategy consistently outperforms the "All in One" approach. While the "Random" assignment benefits the base model LEADS, it slightly diminishes CoDA's performance across all datasets. \model~shows strong adaptation capabilities across various datasets and base generalization methods, consistently outperforming other non-Oracle techniques. This indicates that \model~effectively captures commonalities across environments, enabling smoother adaptation to new conditions. Furthermore, the performance gap between \model~and the Oracle is significantly narrower in adaptation tasks compared to in-domain generalization.
\begin{table*}[htb]
\centering
\renewcommand\arraystretch{1.1}
\resizebox{\linewidth}{!}{
\begin{tabular}{c|c|cc|cc|cc}
\toprule[1.5pt]
\multirow{2}{*}{Data}  & \multirow{2}{*}{Assignment} & \multicolumn{2}{c|}{LEADS} & \multicolumn{2}{c|}{CoDA-$l_1$} & \multicolumn{2}{c}{CoDA-$l_2$}  \\ 
& & MSE & MAPE & MSE & MAPE & MSE & MAPE
\\
\hline \multirow{5}{*}{LV} & All in One  & 4.16±8.61 E-2 & 9.92±3.55 &  4.01±6.43 E-2   & 26.90±10.65 & 4.10±7.61 E-2  & 27.80±8.81  \\

& One per Env   & 2.28±1.81 E-3 & 5.41±0.50 & 1.72±0.53 E-3  & 27.63±7.81  & 1.66±0.82 E-3 & 25.51±7.72   \\

& Random   & 1.72±0.53 E-3 & 6.87±0.13 & 1.14±0.61 E-3    & 27.56±6.36 & 1.05±0.54 E-3 & 29.65±6.31  \\

& \cellcolor{Gray_l} \model  &\cellcolor{Gray_l} \textbf{5.77±1.46 E-4}  & \cellcolor{Gray_l} \textbf{2.84±0.13} & \cellcolor{Gray_l} \textbf{8.37±0.94 E-5}    & \cellcolor{Gray_l} \textbf{10.16±0.04}  &\cellcolor{Gray_l} \textbf{8.49±2.07 E-5}  & \cellcolor{Gray_l} \textbf{10.30±0.08} \\

& \cellcolor{Gray_d} Oracle                  & \cellcolor{Gray_d} 1.67±2.26 E-3 & \cellcolor{Gray_d} 3.16±0.37  &\cellcolor{Gray_d} 5.85±1.24 E-5   &\cellcolor{Gray_d} 10.24±0.06 &\cellcolor{Gray_d} 5.12±3.17 E-5 &\cellcolor{Gray_d} 10.24±0.04 \\ 

\hline \multirow{5}{*}{GS} & All in One    
& 4.59±1.18 E-4 & 721.25±197.54 & 2.85±0.55 E-3 & 6658.20±1651.77  
&2.76±0.72 E-3 & 7355.20±335.45  \\

& One per Env  & 3.59±2.71 E-4 & 450.20±368.36 & 1.08±0.82 E-3 & 6247.34±817.74 & 1.19±0.97 E-3 & 5948.57±935.71    \\

& Random   & 5.73±0.86 E-4 & 1261.00±979.30 & 2.92±0.80 E-3 &  7292.88±312.54 & 2.84±0.89 E-3 & 4499.25±844.62 \\

& \cellcolor{Gray_l} \model   &\cellcolor{Gray_l} \textbf{1.00±0.32 E-4} &\cellcolor{Gray_l} \textbf{378.73±182.12} & \cellcolor{Gray_l}\textbf{2.41±0.91 E-4} & \cellcolor{Gray_l} \textbf{220.54±65.60}
 &\cellcolor{Gray_l} \textbf{2.13±0.41 E-4} 
 &\cellcolor{Gray_l} \textbf{207.96±46.49 }
 \\
 
& \cellcolor{Gray_d} Oracle &\cellcolor{Gray_d} 2.21±0.93 E-4
&\cellcolor{Gray_d} 434.73±432.38 &\cellcolor{Gray_d} 2.66±0.79 E-4 & \cellcolor{Gray_d} 302.68±188.25  &\cellcolor{Gray_d} 2.10±0.89 E-4 &\cellcolor{Gray_d} 230.62±118.29  \\ 

\hline \multirow{5}{*}{NS} & All in One        
& 1.25±2.04 E-2 & 67.17±3.33   
&1.25±0.20 E-2 & 218.66±27.26  
&1.29±0.29 E-2 & 214.44±17.08  \\
& One per Env  
& 2.78±2.08 E-2 & 96.23±4.54
&2.04±0.68 E-2 & 214.38±15.19
&2.43±0.48 E-2 & 209.68±18.98
\\
& Random   
& 1.32±0.53 E-2 & 81.18±7.43
&4.66±1.04 E-2 & 215.21±19.39  
&4.37±0.99 E-2 & 191.03±16.38  \\

& \cellcolor{Gray_l} \model   
& \cellcolor{Gray_l} \textbf{7.52±0.76 E-3}  &\cellcolor{Gray_l} \textbf{50.93±8.83}
&\cellcolor{Gray_l} \textbf{9.27±1.81 E-3} 
&\cellcolor{Gray_l} \textbf{101.38±15.77} 
&\cellcolor{Gray_l} \textbf{9.71±2.10 E-3} 
&\cellcolor{Gray_l} \textbf{101.04±16.27}
\\
& \cellcolor{Gray_d} Oracle                &\cellcolor{Gray_d} 1.16±0.68 E-2
&\cellcolor{Gray_d} 57.35±17.90
&\cellcolor{Gray_d} 7.46±0.72 E-3
&\cellcolor{Gray_d} 154.86±41.00
&\cellcolor{Gray_d} 7.32±0.81 E-3 
&\cellcolor{Gray_d} 100.04±24.93  
\\ 
\bottomrule[1.5pt]
\end{tabular}
}
\caption{Adaptation Experiment Results on the LV, GS, and NS environments.
\model~consistently outperforms other non-Oracle methods across all datasets and narrows the performance gap with the Oracle more effectively compared to in-domain generalization.
}
\label{table_adapt}
\end{table*}

\begin{figure*}[htb]
\centering
\subfloat{\includegraphics[width = 0.9\linewidth, keepaspectratio]{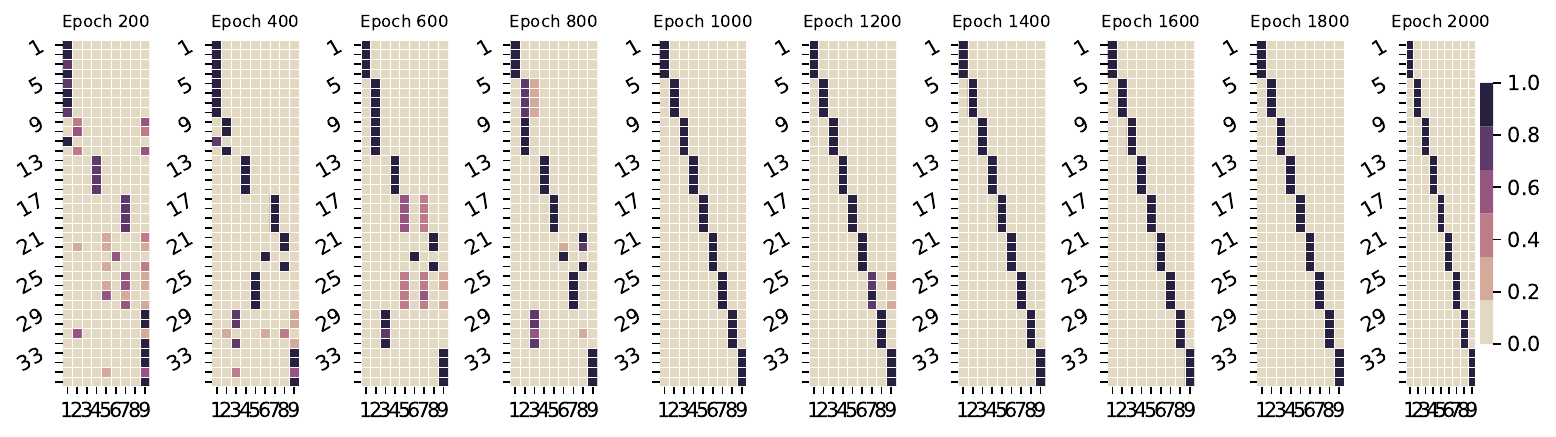}} \\
\vspace{-10pt}
\subfloat{\includegraphics[width = 0.9\linewidth, keepaspectratio]{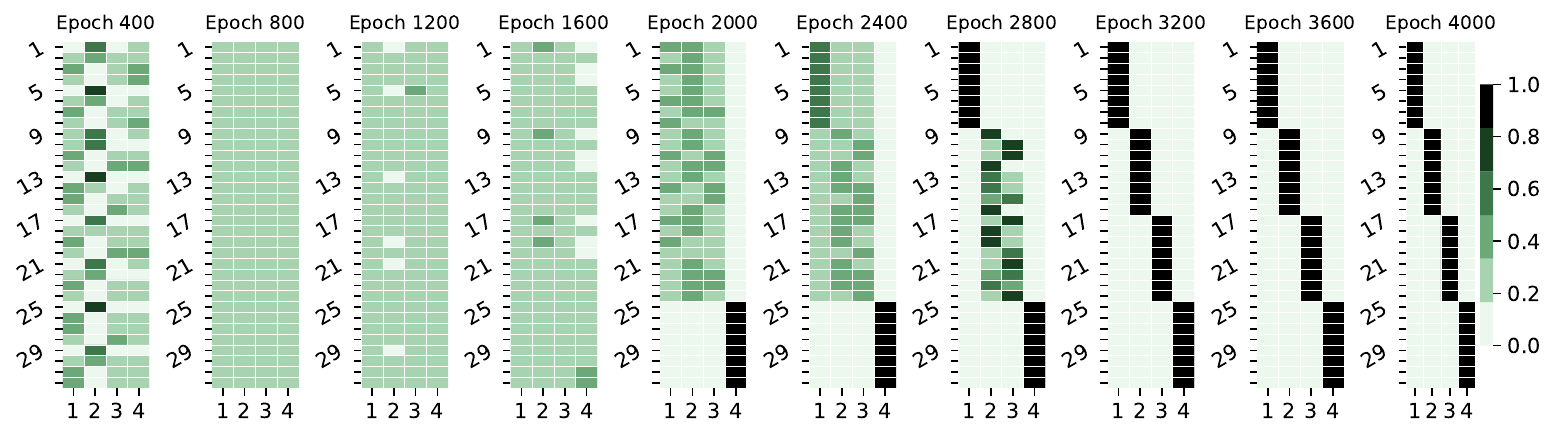}} 
\caption{Environment assignment probability over time, averaged over 5 runs, with LEADS as base model (on LV (top) and NS (bottom); see Appendix \ref{sec:fig_conv_gs} for GS). The assignment converges to the true label faster than the designated training steps. A similar trend is observed with the CoDA model.}
\label{fig_assignment}
\end{figure*}
\begin{figure*}[htb]
\centering
\includegraphics[width=\linewidth]{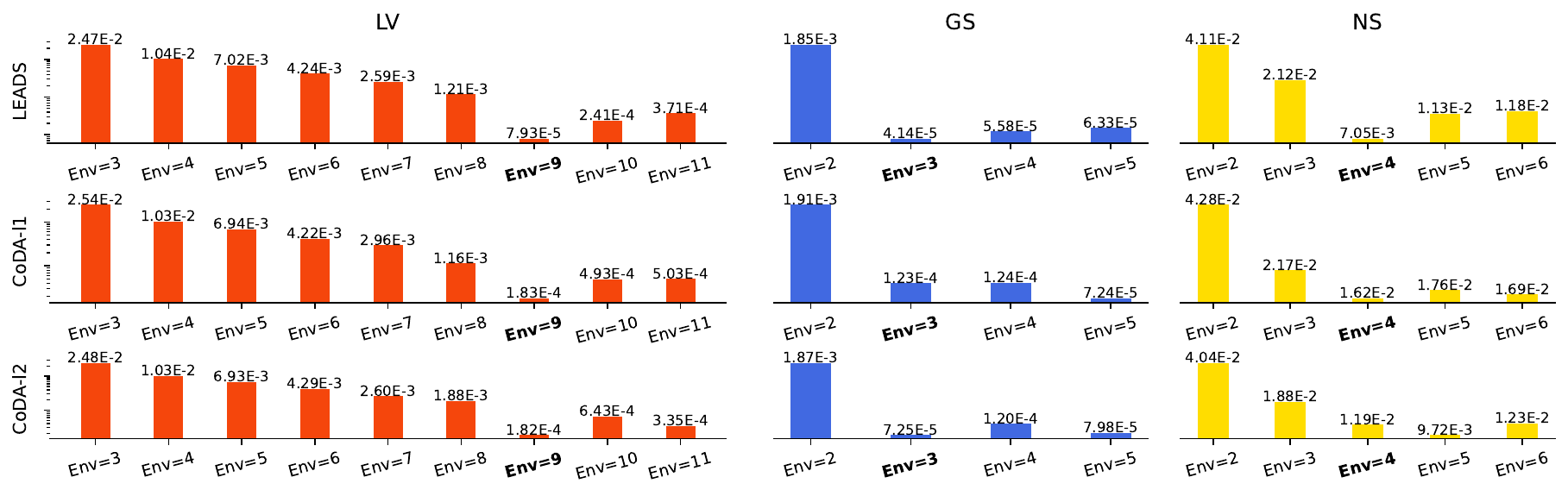}
\caption{Performance across assumed number of environments with \model (log-scale y-axis). The peak performance aligns with the true number of environments (bold on x-axis) with high probability, and remains stable thereafter.}
\label{fig_k}
\end{figure*}

\subsubsection{Assignments Convergence} We illustrate the probability of environment assignments with \model~over training time in Figure~\ref{fig_assignment}. Initially, our model may default to random assignments due to unoptimized neural networks. However, the assignments quickly converge to the true labels. 
Notably, systems with simpler dynamics, like LV compared to NS, enable quicker learning of base generalization methods, resulting in faster convergence of environment assignments.

\subsubsection{Performance across Varying Number of Assumed Environments} 
As the true number of environments ($|\mathcal{E}_o|$) might be unknown, we examine \model's performance with varying assumed environments $M$ in Figure~\ref{fig_k}. Our findings show that prior knowledge of the true $M$ is beneficial: performance peaks when the assumed $M$ aligns with the true count.
Additionally, our model demonstrates robustness to over-estimations of $M$. 
Due to its ability to account for bias, our model effectively identifies trajectories from the same environment and remains robust to an excess of inaccurately trained neural networks, even when the assumed $M$ is too large.
The above observations suggest that incrementally increasing the number of environments until peak performance is achieved can be a straightforward way to identify the true $M$.
Lastly, DynaInfer consistently outperforms other non-oracle approaches when $M$ is underestimated, except for the One-per-Env baseline (which requires $M$ equal to the number of trajectories and is computationally infeasible).

\section{Related Work}

\subsection{Domain Generalization and Adaptation} Domain generalization (DG) seeks to train a model on one or multiple distinct but related source domains so that it generalizes effectively to any out-of-distribution (OOD) target domain. DG methods assume data heterogeneity and use additional environment labels to develop models that remain robust across unseen and shifted test data. Many DG strategies focus on domain alignment, aiming to minimize divergence among source domains to achieve domain-invariant representations~\citep{li2018domain,hu2020domain,piratla2020efficient,seo2020learning,wang2023out}. Other approaches enhance the diversity of training data by augmenting source domains~\citep{carlucci2019domain, shankar2018generalizing, volpi2018generalizing}. Additionally, some methods leverage meta-learning and invariant risk minimization for regularization, further enhancing generalization~\citep{li2019episodic,arjovsky2019invariant}.

Domain adaptation (DA) methods enable model generalization to target domains with shifted data distributions and are primarily classified into three categories.
Instance-based methods reweight or adjust training samples to reflect the test distribution~\citep{DBLP:conf/acl/JiangZ07,DBLP:conf/icml/DaiYXY07}. 
Feature-based approaches align feature distributions across training and test domains~\citep{DBLP:journals/corr/TzengHZSD14,DBLP:conf/eccv/SunS16}.
Model-based strategies focus on developing models that are either robust to domain shifts or specifically tailored for the target domain~\citep{DBLP:conf/iccv/MotiianPAD17,DBLP:journals/jmlr/GaninUAGLLML16}.

\subsection{Generalization for Dynamical Systems}
Generalization in dynamical systems remains underexplored in literature. 
Among the limited studies, LEADS emerges as a novel multi-task learning framework that effectively generalizes across the functional space of dynamical systems~\citep{yin2021leads}.
Alternatively, CoDA optimizes within the parameter space, enhancing model adaptability and efficiency while accommodating increased environmental variability without requiring multiple distinct network trainings for each setting~\citep{kirchmeyer2022generalizing}. In contrast, DyAd is a context-aware meta-learning approach that adjusts the dynamics model by decoding a time-invariant context from observed states~\citep{wang2022meta}. Despite its novelty, DyAd relies on potentially impractical weak supervision based on physics-derived quantities and uses Adaptive Instance Normalization, which may degrade performance.

Currently, three notable weaknesses prevail in generalization works for dynamical systems. First, there is an assumption that prior knowledge about the target domain exists, and without it, most generalization methods would fail~\citep{goring2024out}. Second, the predominant use of the mean squared error as a loss function is inadequate for evaluating the reconstruction accuracy of chaotic systems. Lastly, the influence of unlabelled trajectory data on the process of learning generalizable dynamical systems remains both unexplored and unresolved — a gap this paper examines for the first time.
While switched systems learning methods~\citep{lauer2013estimating} infer modes by classifying individual data points (analogous to environment inference), our work operates on trajectories governed by ODEs or PDEs.


\section{Conclusion}
We propose an environment inference method that improves the understanding and generalization of complex dynamical systems across various environments without using manually labeled data. \model~infers environment labels directly from training data, overcoming the challenges associated with explicit annotation. Theoretical analysis ensures convergence of \model, and experiments show \model~often surpasses non-oracle methods and matches or exceeds oracle performance.

Future research could unfold along the following promising directions: 
First, to improve generalization in chaotic systems, MSE-based methods could be replaced with more suitable metrics such as the sliced Wasserstein-1 distance, which would require developing a tailored inference model.
Similarly, effectively inferring environments for dynamics with complex boundaries remains a significant open problem, as current learning methods often oversimplify boundary conditions.
Furthermore, for systems requiring interpretable parameters, our method could be extended to jointly optimize environment labels and physical coefficients.
To improve convergence, coordinate optimization techniques may help escape local optima for objectives that are convex in individual variable blocks~\citep{helton1997coordinate,2025rome}.
Finally, techniques such as adaptive early stopping, dynamic batching, and membership functions~\citep{bianchi2022constrained} could further enhance training efficiency.



\begin{ack}
This work was supported by NSFC (No. 62425206, 62141607, 62525213), Beijing Municipal Science and Technology Project (No. Z241100004224009), and Big Data and Responsible Artificial Intelligence for National Governance, Renmin University of China.
We also extend our thanks to Renzhe Xu and Hao Zou for their insightful discussions.
\end{ack}

\bibliographystyle{plain}
\bibliography{reference}

\begin{thebibliography}{10}

\bibitem{arjovsky2019invariant}
Martin Arjovsky, L{\'e}on Bottou, Ishaan Gulrajani, and David Lopez-Paz.
\newblock Invariant risk minimization.
\newblock {\em arXiv preprint arXiv:1907.02893}, 2019.

\bibitem{ben2009robust}
Aharon Ben-Tal, Laurent El~Ghaoui, and Arkadi Nemirovski.
\newblock {\em Robust optimization}, volume~28.
\newblock Princeton university press, 2009.

\bibitem{bianchi2022constrained}
Federico Bianchi, Alessandro Falsone, Luigi Piroddi, and Maria Prandini.
\newblock A constrained clustering approach to bounded-error identification of switched and piecewise affine systems.
\newblock {\em Automatica}, 146:110589, 2022.

\bibitem{brunton2016discovering}
Steven~L Brunton, Joshua~L Proctor, and J~Nathan Kutz.
\newblock Discovering governing equations from data by sparse identification of nonlinear dynamical systems.
\newblock {\em Proceedings of the national academy of sciences}, 113(15):3932--3937, 2016.

\bibitem{carlucci2019domain}
Fabio~M Carlucci, Antonio D'Innocente, Silvia Bucci, Barbara Caputo, and Tatiana Tommasi.
\newblock Domain generalization by solving jigsaw puzzles.
\newblock In {\em Proceedings of the IEEE/CVF Conference on Computer Vision and Pattern Recognition}, pages 2229--2238, 2019.

\bibitem{DBLP:conf/icml/DaiYXY07}
Wenyuan Dai, Qiang Yang, Gui{-}Rong Xue, and Yong Yu.
\newblock Boosting for transfer learning.
\newblock In Zoubin Ghahramani, editor, {\em Machine Learning, Proceedings of the Twenty-Fourth International Conference {(ICML} 2007), Corvallis, Oregon, USA, June 20-24, 2007}, volume 227 of {\em {ACM} International Conference Proceeding Series}, pages 193--200. {ACM}, 2007.

\bibitem{de2019deep}
Emmanuel De~B{\'e}zenac, Arthur Pajot, and Patrick Gallinari.
\newblock Deep learning for physical processes: Incorporating prior scientific knowledge.
\newblock {\em Journal of Statistical Mechanics: Theory and Experiment}, 2019(12):124009, 2019.

\bibitem{dormand1980family}
John~R Dormand and Peter~J Prince.
\newblock A family of embedded runge-kutta formulae.
\newblock {\em Journal of computational and applied mathematics}, 6(1):19--26, 1980.

\bibitem{duraisamy2019turbulence}
Karthik Duraisamy, Gianluca Iaccarino, and Heng Xiao.
\newblock Turbulence modeling in the age of data.
\newblock {\em Annual review of fluid mechanics}, 51:357--377, 2019.

\bibitem{DBLP:journals/jmlr/GaninUAGLLML16}
Yaroslav Ganin, Evgeniya Ustinova, Hana Ajakan, Pascal Germain, Hugo Larochelle, Fran{\c{c}}ois Laviolette, Mario Marchand, and Victor~S. Lempitsky.
\newblock Domain-adversarial training of neural networks.
\newblock {\em J. Mach. Learn. Res.}, 17:59:1--59:35, 2016.

\bibitem{goring2024out}
Niclas G{\"o}ring, Florian Hess, Manuel Brenner, Zahra Monfared, and Daniel Durstewitz.
\newblock Out-of-domain generalization in dynamical systems reconstruction.
\newblock In {\em Proceedings of the 41st International Conference on Machine Learning}, pages 16071--16114, 2024.

\bibitem{harris2020array}
Charles~R Harris, K~Jarrod Millman, St{\'e}fan~J Van Der~Walt, Ralf Gommers, Pauli Virtanen, David Cournapeau, Eric Wieser, Julian Taylor, Sebastian Berg, Nathaniel~J Smith, et~al.
\newblock Array programming with numpy.
\newblock {\em Nature}, 585(7825):357--362, 2020.

\bibitem{helton1997coordinate}
J~William Helton and Orlando Merino.
\newblock Coordinate optimization for bi-convex matrix inequalities.
\newblock In {\em Proceedings of the 36th IEEE Conference on Decision and Control}, volume~4, pages 3609--3613. IEEE, 1997.

\bibitem{hu2020domain}
Shoubo Hu, Kun Zhang, Zhitang Chen, and Laiwan Chan.
\newblock Domain generalization via multidomain discriminant analysis.
\newblock In {\em Uncertainty in Artificial Intelligence}, pages 292--302. PMLR, 2020.

\bibitem{DBLP:conf/acl/JiangZ07}
Jing Jiang and ChengXiang Zhai.
\newblock Instance weighting for domain adaptation in {NLP}.
\newblock In John Carroll, Antal van~den Bosch, and Annie Zaenen, editors, {\em {ACL} 2007, Proceedings of the 45th Annual Meeting of the Association for Computational Linguistics, June 23-30, 2007, Prague, Czech Republic}. The Association for Computational Linguistics, 2007.

\bibitem{khansari2011learning}
S~Mohammad Khansari-Zadeh and Aude Billard.
\newblock Learning stable nonlinear dynamical systems with gaussian mixture models.
\newblock {\em IEEE Transactions on Robotics}, 27(5):943--957, 2011.

\bibitem{kirchmeyer2022generalizing}
Matthieu Kirchmeyer, Yuan Yin, J{\'e}r{\'e}mie Don{\`a}, Nicolas Baskiotis, Alain Rakotomamonjy, and Patrick Gallinari.
\newblock Generalizing to new physical systems via context-informed dynamics model.
\newblock In {\em International Conference on Machine Learning}, pages 11283--11301. PMLR, 2022.

\bibitem{lahoti2020fairness}
Preethi Lahoti, Alex Beutel, Jilin Chen, Kang Lee, Flavien Prost, Nithum Thain, Xuezhi Wang, and Ed~Chi.
\newblock Fairness without demographics through adversarially reweighted learning.
\newblock {\em Advances in neural information processing systems}, 33:728--740, 2020.

\bibitem{lauer2013estimating}
Fabien Lauer.
\newblock Estimating the probability of success of a simple algorithm for switched linear regression.
\newblock {\em Nonlinear Analysis: Hybrid Systems}, 8:31--47, 2013.

\bibitem{li2019episodic}
Da~Li, Jianshu Zhang, Yongxin Yang, Cong Liu, Yi-Zhe Song, and Timothy~M Hospedales.
\newblock Episodic training for domain generalization.
\newblock In {\em Proceedings of the IEEE/CVF International Conference on Computer Vision}, pages 1446--1455, 2019.

\bibitem{li2018domain}
Haoliang Li, Sinno~Jialin Pan, Shiqi Wang, and Alex~C Kot.
\newblock Domain generalization with adversarial feature learning.
\newblock In {\em Proceedings of the IEEE conference on computer vision and pattern recognition}, pages 5400--5409, 2018.

\bibitem{li2020fourier}
Zongyi Li, Nikola~Borislavov Kovachki, Kamyar Azizzadenesheli, Kaushik Bhattacharya, Andrew Stuart, Anima Anandkumar, et~al.
\newblock Fourier neural operator for parametric partial differential equations.
\newblock In {\em International Conference on Learning Representations}.

\bibitem{lotka1925elements}
Alfred~James Lotka.
\newblock {\em Elements of physical biology}.
\newblock Williams \& Wilkins, 1925.

\bibitem{madec2017nemo}
Gurvan Madec, Romain Bourdall{\'e}-Badie, Pierre-Antoine Bouttier, Cl{\'e}ment Bricaud, Diego Bruciaferri, Daley Calvert, J{\'e}r{\^o}me Chanut, Emanuela Clementi, Andrew Coward, Damiano Delrosso, et~al.
\newblock Nemo ocean engine.
\newblock 2017.

\bibitem{DBLP:conf/iccv/MotiianPAD17}
Saeid Motiian, Marco Piccirilli, Donald~A. Adjeroh, and Gianfranco Doretto.
\newblock Unified deep supervised domain adaptation and generalization.
\newblock In {\em {IEEE} International Conference on Computer Vision, {ICCV} 2017, Venice, Italy, October 22-29, 2017}, pages 5716--5726. {IEEE} Computer Society, 2017.

\bibitem{neic2017efficient}
Aurel Neic, Fernando~O Campos, Anton~J Prassl, Steven~A Niederer, Martin~J Bishop, Edward~J Vigmond, and Gernot Plank.
\newblock Efficient computation of electrograms and ecgs in human whole heart simulations using a reaction-eikonal model.
\newblock {\em Journal of computational physics}, 346:191--211, 2017.

\bibitem{paszke2019pytorch}
Adam Paszke, Sam Gross, Francisco Massa, Adam Lerer, James Bradbury, Gregory Chanan, Trevor Killeen, Zeming Lin, Natalia Gimelshein, Luca Antiga, et~al.
\newblock Pytorch: An imperative style, high-performance deep learning library.
\newblock {\em Advances in neural information processing systems}, 32, 2019.

\bibitem{pearson1993complex}
John~E Pearson.
\newblock Complex patterns in a simple system.
\newblock {\em Science}, 261(5118):189--192, 1993.

\bibitem{piratla2020efficient}
Vihari Piratla, Praneeth Netrapalli, and Sunita Sarawagi.
\newblock Efficient domain generalization via common-specific low-rank decomposition.
\newblock In {\em International Conference on Machine Learning}, pages 7728--7738. PMLR, 2020.

\bibitem{2025rome}
Tianle Pu, Zijie Geng, Haoyang Liu, Shixuan Liu, Jie Wang, Li~Zeng, Chao Chen, and Changjun Fan.
\newblock {RoME}: Domain-robust mixture-of-experts for {MILP} solution prediction across domains.
\newblock In {\em Advances in Neural Information Processing Systems}, volume~39, 2025.

\bibitem{seo2020learning}
Seonguk Seo, Yumin Suh, Dongwan Kim, Geeho Kim, Jongwoo Han, and Bohyung Han.
\newblock Learning to optimize domain specific normalization for domain generalization.
\newblock In {\em Computer Vision--ECCV 2020: 16th European Conference, Glasgow, UK, August 23--28, 2020, Proceedings, Part XXII 16}, pages 68--83. Springer, 2020.

\bibitem{shaier2021data}
Sagi Shaier, Maziar Raissi, and Padmanabhan Seshaiyer.
\newblock Data-driven approaches for predicting spread of infectious diseases through dinns: Disease informed neural networks.
\newblock {\em Letters in Biomathematics}, 9(1):71--105, 2022.

\bibitem{shankar2018generalizing}
Shiv Shankar, Vihari Piratla, Soumen Chakrabarti, Siddhartha Chaudhuri, Preethi Jyothi, and Sunita Sarawagi.
\newblock Generalizing across domains via cross-gradient training.
\newblock In {\em International Conference on Learning Representations}, 2018.

\bibitem{sirignano2018dgm}
Justin Sirignano and Konstantinos Spiliopoulos.
\newblock Dgm: A deep learning algorithm for solving partial differential equations.
\newblock {\em Journal of computational physics}, 375:1339--1364, 2018.

\bibitem{srivastava2020robustness}
Megha Srivastava, Tatsunori Hashimoto, and Percy Liang.
\newblock Robustness to spurious correlations via human annotations.
\newblock In {\em International Conference on Machine Learning}, pages 9109--9119. PMLR, 2020.

\bibitem{DBLP:conf/eccv/SunS16}
Baochen Sun and Kate Saenko.
\newblock Deep {CORAL:} correlation alignment for deep domain adaptation.
\newblock In Gang Hua and Herv{\'{e}} J{\'{e}}gou, editors, {\em Computer Vision - {ECCV} 2016 Workshops - Amsterdam, The Netherlands, October 8-10 and 15-16, 2016, Proceedings, Part {III}}, volume 9915 of {\em Lecture Notes in Computer Science}, pages 443--450, 2016.

\bibitem{sun2012regularized}
Wei Sun, Junhui Wang, and Yixin Fang.
\newblock Regularized k-means clustering of high-dimensional data and its asymptotic consistency.
\newblock {\em Electronic Journal of Statistics}, 6:148--167, 2012.

\bibitem{DBLP:journals/corr/TzengHZSD14}
Eric Tzeng, Judy Hoffman, Ning Zhang, Kate Saenko, and Trevor Darrell.
\newblock Deep domain confusion: Maximizing for domain invariance.
\newblock {\em arXiv preprint arXiv:1412.3474}, 2014.

\bibitem{volpi2018generalizing}
Riccardo Volpi, Hongseok Namkoong, Ozan Sener, John~C Duchi, Vittorio Murino, and Silvio Savarese.
\newblock Generalizing to unseen domains via adversarial data augmentation.
\newblock {\em Advances in neural information processing systems}, 31, 2018.

\bibitem{wang2023out}
Haotian Wang, Kun Kuang, Long Lan, Zige Wang, Wanrong Huang, Fei Wu, and Wenjing Yang.
\newblock Out-of-distribution generalization with causal feature separation.
\newblock {\em IEEE Transactions on Knowledge and Data Engineering}, 36(4):1758--1772, 2023.

\bibitem{wang2022meta}
Rui Wang, Robin Walters, and Rose Yu.
\newblock Meta-learning dynamics forecasting using task inference.
\newblock {\em Advances in Neural Information Processing Systems}, 35:21640--21653, 2022.

\bibitem{willard2020integrating}
Jared Willard, Xiaowei Jia, Shaoming Xu, Michael Steinbach, and Vipin Kumar.
\newblock Integrating physics-based modeling with machine learning: A survey.
\newblock {\em arXiv preprint arXiv:2003.04919}, 1(1):1--34, 2020.

\bibitem{yin2021leads}
Yuan Yin, Ibrahim Ayed, Emmanuel de~B{\'e}zenac, Nicolas Baskiotis, and Patrick Gallinari.
\newblock Leads: Learning dynamical systems that generalize across environments.
\newblock {\em Advances in Neural Information Processing Systems}, 34:7561--7573, 2021.

\bibitem{zhi2022learning}
Weiming Zhi, Tin Lai, Lionel Ott, Edwin~V Bonilla, and Fabio Ramos.
\newblock Learning efficient and robust ordinary differential equations via invertible neural networks.
\newblock In {\em International Conference on Machine Learning}, pages 27060--27074. PMLR, 2022.

\end{thebibliography}

\clearpage
\section*{NeurIPS Paper Checklist}

\begin{enumerate}

\item {\bf Claims}
    \item[] Question: Do the main claims made in the abstract and introduction accurately reflect the paper's contributions and scope?
    \item[] Answer: \answerYes{} 
    \item[] Justification: The abstract and introduction accurately outline our research questions, and faithfully reflect the paper’s contributions and scope.

\item {\bf Limitations}
    \item[] Question: Does the paper discuss the limitations of the work performed by the authors?
    \item[] Answer: \answerYes{} 
    \item[] Justification: We discussed it as future work in the Conclusion section

\item {\bf Theory Assumptions and Proofs}
    \item[] Question: For each theoretical result, does the paper provide the full set of assumptions and a complete (and correct) proof?
    \item[] Answer: \answerYes{} 
    \item[] Justification: Assumptions, formal statements of theories and proofs can be found in appendix.

    \item {\bf Experimental Result Reproducibility}
    \item[] Question: Does the paper fully disclose all the information needed to reproduce the main experimental results of the paper to the extent that it affects the main claims and/or conclusions of the paper (regardless of whether the code and data are provided or not)?
    \item[] Answer: \answerYes{} 
    \item[] Justification: Experiment settings and implementation of methods are described in the experiment section. Codes are available in supplementary materials.

\item {\bf Open access to data and code}
    \item[] Question: Does the paper provide open access to the data and code, with sufficient instructions to faithfully reproduce the main experimental results, as described in supplemental material?
    \item[] Answer: \answerYes{} 
    \item[] Justification: All the datasets in this paper are public with citations (see section 6.1). Code is provided in additional supplemental materials.

\item {\bf Experimental Setting/Details}
    \item[] Question: Does the paper specify all the training and test details (e.g., data splits, hyperparameters, how they were chosen, type of optimizer, etc.) necessary to understand the results?
    \item[] Answer: \answerYes{} 
    \item[] Justification: Experiment settings are described in section 4.1 and 4.2.

\item {\bf Experiment Statistical Significance}
    \item[] Question: Does the paper report error bars suitably and correctly defined or other appropriate information about the statistical significance of the experiments?
    \item[] Answer: \answerYes{} 
    \item[] Justification: We use mean $\pm$ standard deviation to report the results over 5 independent runs. See section 4.2 for details.

\item {\bf Experiments Compute Resources}
    \item[] Question: For each experiment, does the paper provide sufficient information on the computer resources (type of compute workers, memory, time of execution) needed to reproduce the experiments?
    \item[] Answer: \answerYes{} 
    \item[] Justification: Refer to Appendix~\ref{appendix_environments}. Results are averaged over 5 independent runs.
    
\item {\bf Code Of Ethics}
    \item[] Question: Does the research conducted in the paper conform, in every respect, with the NeurIPS Code of Ethics \url{https://neurips.cc/public/EthicsGuidelines}?
    \item[] Answer: \answerYes{} 
    \item[] Justification: The research conducted in this paper conforms with the NeurIPS Code of Ethics

\item {\bf Broader Impacts}
    \item[] Question: Does the paper discuss both potential positive societal impacts and negative societal impacts of the work performed?
    \item[] Answer: \answerNA{} 
    \item[] Justification: There are no negative social impacts for research conducted in this paper.
    
\item {\bf Safeguards}
    \item[] Question: Does the paper describe safeguards that have been put in place for responsible release of data or models that have a high risk for misuse (e.g., pretrained language models, image generators, or scraped datasets)?
    \item[] Answer: \answerNA{} 
    \item[] Justification: The paper poses no safeguards risks.

\item {\bf Licenses for existing assets}
    \item[] Question: Are the creators or original owners of assets (e.g., code, data, models), used in the paper, properly credited and are the license and terms of use explicitly mentioned and properly respected?
    \item[] Answer: \answerYes{} 
    \item[] Justification: All datasets and base models in this paper are cited. Codes are credited with original licenses provided.

\item {\bf New Assets}
    \item[] Question: Are new assets introduced in the paper well documented and is the documentation provided alongside the assets?
    \item[] Answer: \answerNA{} 
    \item[] Justification: The paper does not release new assets.

\item {\bf Crowdsourcing and Research with Human Subjects}
    \item[] Question: For crowdsourcing experiments and research with human subjects, does the paper include the full text of instructions given to participants and screenshots, if applicable, as well as details about compensation (if any)? 
    \item[] Answer: \answerNA{} 
    \item[] Justification: The paper does not involve crowdsourcing nor research with human subjects

\item {\bf Institutional Review Board (IRB) Approvals or Equivalent for Research with Human Subjects}
    \item[] Question: Does the paper describe potential risks incurred by study participants, whether such risks were disclosed to the subjects, and whether Institutional Review Board (IRB) approvals (or an equivalent approval/review based on the requirements of your country or institution) were obtained?
    \item[] Answer: \answerNA{} 
    \item[] Justification: The paper does not involve crowd-sourcing nor research with human subjects.

    \item {\bf Declaration of LLM usage}
    \item[] Question: Does the paper describe the usage of LLMs if it is an important, original, or non-standard component of the core methods in this research? Note that if the LLM is used only for writing, editing, or formatting purposes and does not impact the core methodology, scientific rigorousness, or originality of the research, declaration is not required.
    \item[] Answer: \answerNA{} 
    \item[] Justification: The core method development in this research does not involve LLMs.

\end{enumerate}

\newpage
\appendix
\onecolumn

\section{Proof of Proposition~\ref{prop:convergence}} \label{sec:proof-convergence}
\begin{proof}
Due to the operation in Equation~\eqref{eq:theta-phi-updating}, we must have
$$
R_{\hat{\bfe}^{(r+1)}}\left(\theta^{(r+1)}, \bfphi^{(r+1)}\right) \le R_{\hat{\bfe}^{(r+1)}}\left(\theta^{(r)}, \bfphi^{(r)}\right).
$$
In addition, according to the operation in Equation~\eqref{eq:e-updating}, we must have
\begin{equation*}
\begin{split}
\forall i \in [N], & \int_{t\in I} \left\|\frac{dx^{i}_t}{dt} - h\left(x^{i}_t; \theta^{(r)}, \phi_{e_i^{(r+1)}}^{(r)}\right)\right\|_2^2dt \\
&\le \int_{t\in I} \left\|\frac{dx^{i}_t}{dt} - h\left(x^{i}_t; \theta^{(r)}, \phi_{e_i^{(r)}}^{(r)}\right)\right\|_2^2dt.
\end{split}
\end{equation*}

As a result, since the regularization terms (\textit{i.e.}, the $\Omega(\phi_e)$ term) in $R_{\hat{\bfe^{r+1}}}(\theta^{(r)}, \bfphi^{(r)})$ and $R_{\hat{\bfe^{r}}}(\theta^{(r)}, \bfphi^{(r)})$ remain the same, we must have
$$
R_{\hat{\bfe}^{(r+1)}}\left(\theta^{(r)}, \bfphi^{(r)}\right) \le R_{\hat{\bfe}^{(r)}}\left(\theta^{(r)}, \bfphi^{(r)}\right).
$$
Now we have
$$
R_{\hat{\bfe}^{(r+1)}}\left(\theta^{(r+1)}, \bfphi^{(r+1)}\right) \le R_{\hat{\bfe}^{(r+1)}}\left(\theta^{(r)}, \bfphi^{(r)}\right) \le R_{\hat{\bfe}^{(r)}}\left(\theta^{(r)}, \bfphi^{(r)}\right).
$$

Now consider the second part of the proposition. Define the following space of $\theta, \bfphi$
$$
\calH \triangleq \left\{(\theta, \bfphi): \exists \hat{\bfe} \in [M]^N \text{ such that } R_{\hat{\bfe}}(\theta, \bfphi) = \min_{\theta', \bfphi'} R_{\hat{\bfe}}(\theta', \bfphi')\right\}.
$$
Note that by the assumption, we must have $|\calH| < \infty$. Now define $\calA$ as
$$
\calA = \left\{\int_{t\in I} \left\|\frac{dx^{i}_t}{dt} - h\left(x^{i}_t; \theta, \phi_{e}\right)\right\|_2^2dt: i\in [N], e \in [M], (\theta, \bfphi) \in \calH\right\}.
$$
Note that since $|\calH| < \infty$, we have $|\calA| < \infty$. $C$ is defined as
$$
C = \min_{a, b \in \calA, a \ne b} |a - b|.
$$
Since $|\calA| < \infty$, we must have $C > 0$. In addition, note that when $r > 1$ and $R_{\hat{\bfe}^{(r + 1)}}(\theta^{(r + 1)}, \bfphi^{(r + 1)}) < R_{\hat{\bfe}^{(r)}}( \theta^{(r)}, \bfphi^{(r)})$, we must have $\hat{\bfe}^{(r+1)} \ne \hat{\bfe}^{(r)}$. Otherwise we will have
\begin{equation*}
\begin{split}
R_{\hat{\bfe}^{(r + 1)}}\left(\theta^{(r + 1)}, \bfphi^{(r + 1)}\right) &= R_{\hat{\bfe}^{(r)}}\left(\theta^{(r+1)}, \bfphi^{(r+1)}\right) \\&= R_{\hat{\bfe}^{(r)}}\left(\theta^{(r)}, \bfphi^{(r)}\right).
\end{split}
\end{equation*}
As a result, there exists $i \in [N]$ such that $\hat{e}^{(r+1)}_i \ne \hat{e}^{(r)}_i$. By the choice of $\hat{e}^{(r+1)}_i$, we must have
\begin{equation*}
\begin{split}
\int_{t\in I} & \left\|\frac{dx^{i}_t}{dt} - h\left(x^{i}_t; \theta^{(r)}, \phi^{(r)}_{\hat{e}_i^{(r)}}\right)\right\|_2^2dt \\& \ne \int_{t\in I} \left\|\frac{dx^{i}_t}{dt} - h\left(x^{i}_t; \theta^{(r)}, \phi^{(r)}_{\hat{e}_i^{(r+1)}}\right)\right\|_2^2dt.
\end{split}
\end{equation*}
As a result, by the definition of $C$, we have
\begin{equation*}
\begin{split}
\int_{t\in I} & \left\|\frac{dx^{i}_t}{dt} - h\left(x^{i}_t; \theta^{(r)}, \phi^{(r)}_{\hat{e}_i^{(r)}}\right)\right\|_2^2dt \\& - \int_{t\in I} \left\|\frac{dx^{i}_t}{dt} - h\left(x^{i}_t; \theta^{(r)}, \phi^{(r)}_{\hat{e}_i^{(r+1)}}\right)\right\|_2^2dt \ge C.
\end{split}
\end{equation*}
Therefore, we must have
$$
R_{\hat{\bfe}^{(r + 1)}}\left(\theta^{(r + 1)}, \bfphi^{(r + 1)}\right) \le R_{\hat{\bfe}^{(r)}}\left( \theta^{(r)}, \bfphi^{(r)}\right) - C.
$$
Now the claim follows.
\end{proof}

\section{Real-world Experiment}
We further evaluated our method using a real-world robot motion trajectory dataset \citep{khansari2011learning}. This dataset comprises three distinct motion patterns: (1) drawing "S" shapes, (2) placing a cube on a shelf, and (3) drawing large "C" shapes. In our experimental framework, each of these patterns is treated as a distinct environment. The objective of this evaluation is to assess whether our method can accurately infer the underlying environment, thereby supporting the learning of a generalizable neural network for simulating their dynamics.
To ensure a consistent input structure across environments, all trajectories were projected into a two-dimensional space. The dataset was partitioned into training and test sets with a ratio of 6:1. Furthermore, each trajectory was temporally resampled to a standardized length of 100 time steps for smoothing and uniformity following the practice in~\citep{zhi2022learning}.

The empirical results are presented in Table~\ref{tab:results_rm}. In summary, these findings collectively suggest that our approach maintains its efficacy on real-world data, thereby substantiating its practical utility and robustness.

\begin{table*}[htb]
\centering
\renewcommand\arraystretch{1.1}
\resizebox{\linewidth}{!}{
\begin{tabular}{c|c|ccc|ccc|ccc}
\toprule[1.5pt]
\multirow{3}{*}{Data}  & \multirow{3}{*}{Assignment} & \multicolumn{3}{c|}{LEADS} & \multicolumn{3}{c|}{CoDA-$l_1$} & \multicolumn{3}{c}{CoDA-$l_2$} \\ \cline{3-11}

& & Train     & \multicolumn{2}{c|}{Test}  & Train & \multicolumn{2}{c|}{Test}   & Train       & \multicolumn{2}{c}{Test}    \\ 
& & MSE& MSE & MAPE &MSE & MSE & MAPE &MSE & MSE & MAPE
\\
\hline \multirow{5}{*}{RM} & All in One              & 7.17 E-2  & 7.41±0.02 E-2 & 49.22±1.84 & 7.14 E-2    & 7.40±0.01 E-2   &49.44±3.15   & 7.17 E-2    & 7.41±0.00 E-2  & 39.26±22.13  \\
& One per   Env           & 4.15 E-4  & 4.91±3.50 E-4 & 6.68±2.44 & 8.68 E-4    & 9.14±0.41 E-4  &5.67±1.01  & 8.18 E-4    & 8.43±0.39 E-4 &5.73±1.19   \\
& Random                  & 7.20 E-2  & 7.38±0.02 E-2 & 50.01±1.05 & 7.12 E-2    & 7.39±0.01 E-2    &48.87±1.81 & 7.09 E-2    & 7.39±0.00 E-2  &48.86±2.54  \\
& \cellcolor{Gray_l} \model  & \cellcolor{Gray_l} \textbf{4.74 E-5}  & \cellcolor{Gray_l} \textbf{7.93±2.49 E-5} & \cellcolor{Gray_l} \textbf{2.83±1.62} & \cellcolor{Gray_l} \textbf{9.57 E-5}    & \cellcolor{Gray_l} \textbf{1.83±3.40 E-4}  & \cellcolor{Gray_l} \textbf{3.27±2.36}  & \cellcolor{Gray_l} \textbf{1.71 E-4}    & \cellcolor{Gray_l} \textbf{1.82±3.07 E-4} & \cellcolor{Gray_l} \textbf{2.02±1.66}   \\
& \cellcolor{Gray_d} Oracle                  & \cellcolor{Gray_d} 4.55 E-5  & \cellcolor{Gray_d} 7.02±0.76 E-5 & \cellcolor{Gray_d} 1.78±0.10 &\cellcolor{Gray_d} 1.78 E-5    &\cellcolor{Gray_d} 3.19±0.24 E-5    &\cellcolor{Gray_d} 1.26±0.06 &\cellcolor{Gray_d} 1.99 E-5    &\cellcolor{Gray_d} 2.72±0.18 E-5 &\cellcolor{Gray_d} 1.21±0.08 
\\ 
\bottomrule[1.5pt]
\end{tabular}
}
\caption{In-domain Experiment Results on the Robot Motion.
}
\label{tab:results_rm}
\end{table*}

\section{Sensitivity Analysis to $|\mathcal{E}_o|$}

To assess the robustness of our method to the number of underlying environments, we conducted a sensitivity analysis using the LV system. We systematically varied the true number of environments, $|\mathcal{E}_o|$, from 2 to 16 and evaluated the performance of our model against the Oracle baseline that has privileged access to the true environment labels.
The results are presented in Table~\ref{tab:comparison}. DynaInfer, achieves performance comparable to the Oracle across all tested values of $|\mathcal{E}_o|$. This empirical evidence demonstrates that our approach is robust to variations in the quantity of environments.


\begin{table}[htbp]
\centering
\caption{Test Mean Squared Error on the LV system for varying $|\mathcal{E}_o|$}
\label{tab:comparison}
\resizebox{0.8\linewidth}{!}{
\begin{tabular}{c*{7}{c}}
\toprule[1.5pt]
 & \multicolumn{7}{c}{$\mathcal{E}_o$} \\
\cmidrule(lr){2-8}
 & 2 & 3 & 4 & 5 & 6 & 7 & 8 \\
\midrule
\model & 2.12E-5 & 4.22E-5 & 6.77E-5 & 6.24E-5 & 8.81E-5 & 6.70E-5 & 6.54E-5 \\
Oracle & 2.76E-5 & 3.99E-5 & 5.75E-5 & 5.20E-5 & 7.24E-5 & 8.14E-5 & 5.74E-5 \\
\bottomrule[1.5pt]
\end{tabular}
}
\resizebox{0.9\linewidth}{!}{
\begin{tabular}{c*{8}{c}}
\toprule[1.5pt]
 & \multicolumn{8}{c}{$\mathcal{E}_o$} \\
\cmidrule(lr){2-9}
 & 9 & 10 & 11 & 12 & 13 & 14 & 15 & 16 \\
\midrule
\model & 6.33E-5 & 8.34E-5 & 2.24E-4 & 1.34E-4 & 9.69E-5 & 1.97E-4 & 3.43E-4 & 2.73E-4 \\
Oracle & 7.60E-5 & 9.31E-5 & 1.85E-4 & 1.23E-4 & 1.21E-4 & 1.71E-4 & 2.56E-4 & 2.48E-4 \\
\bottomrule[1.5pt]
\end{tabular}
}
\end{table}

\section{Robustness over $M$}

We assessed our robustness to the overestimation of $M$, the assumed number of environments, in the LV system. The system's true value is $M^* = 9$, with each environment containing four trajectories. In our experiments, we varied the assumed $M$ from 5 to 36. The results, shown in Table~\ref{tab:labels}, demonstrate that \model~effectively adapts to this overestimation. The learned label count consistently converged to a range of 8–11, which aligns closely with the ground-truth value.

Furthermore, we conducted a comprehensive analysis by sweeping $M$ across its full spectrum, from $M=1$ to $M=\mbox{\#trajectory}$. As shown in Table~\ref{tab:mse_m}, the optimal performance was achieved at $M=9$ and the model maintained robustness even when $M$ exceeded this value.

\begin{table}[htbp]
\centering
\caption{Learned Label Counts for different values of $M$ averaged over 5 runs}
\label{tab:labels}
\resizebox{0.95\linewidth}{!}{
\begin{tabular}{c*{16}{c}}
\toprule[1.5pt]
M & 5 & 6 & 7 & 8 & 9 & 10 & 11 & 12 & 13 & 14 & 15 & 16 & 17 & 18 & 19 & 20 \\
\hline
\# Labels & 5 & 6 & 7 & 8 & 9 & 8.8 & 9 & 10.2 & 9 & 9 & 10 & 10.8 & 10 & 11 & 11 & 10 \\
\midrule[1.5pt]
M & 21 & 22 & 23 & 24 & 25 & 26 & 27 & 28 & 29 & 30 & 31 & 32 & 33 & 34 & 35 & 36 \\
\hline
\# Labels & 10.8 & 11 & 9 & 10 & 10.2 & 11 & 10 & 10 & 10 & 11.2 & 10 & 9.2 & 10 & 10.8 & 10 & 11 \\
\bottomrule[1.5pt]
\end{tabular}}
\end{table}

\begin{table}[htbp]
\centering
\caption{Test MSE over the full spectrum of $M$ for the LV system}
\label{tab:mse_m}
\resizebox{0.65\linewidth}{!}{
\begin{tabular}{cccccccc}
\toprule[1.5pt]
$M$ & MSE & $M$ & MSE & $M$ & MSE & $M$ & MSE \\
\midrule
1 & 7.41E-2 & 10 & 2.41E-4 & 19 & 2.60E-4 & 28 & 2.04E-4 \\
2 & 4.01E-2 & 11 & 3.71E-4 & 20 & 1.36E-4 & 29 & 3.65E-4 \\
3 & 2.47E-2 & 12 & 2.57E-4 & 21 & 1.21E-4 & 30 & 1.42E-3 \\
4 & 1.04E-2 & 13 & 1.63E-4 & 22 & 2.45E-4 & 31 & 3.84E-4 \\
5 & 7.02E-3 & 14 & 2.24E-4 & 23 & 9.06E-5 & 32 & 1.16E-4 \\
6 & 4.24E-3 & 15 & 1.15E-4 & 24 & 2.38E-4 & 33 & 3.89E-4 \\
7 & 2.59E-3 & 16 & 1.76E-4 & 25 & 8.27E-5 & 34 & 3.10E-4 \\
8 & 1.21E-3 & 17 & 1.83E-4 & 26 & 1.06E-4 & 35 & 1.72E-4 \\
9 & 7.93E-5 & 18 & 3.60E-4 & 27 & 3.71E-4 & 36 & 2.77E-4 \\
\bottomrule[1.5pt]
\end{tabular}
}
\end{table}

\section{Environment Specification}
We conducted experiments on a server equipped with a 64-core CPU, 256 GB of RAM, and eight 24GB RTX-3090Ti GPUs. The \model~framework was implemented using PyTorch~\citep{paszke2019pytorch}. All NN params in our method are randomly initialized.

\paragraph{Lotka-Volterra (LV)~\citep{lotka1925elements}} The system models the dynamics between a prey-predator pair in an ecosystem, captured by the following ODE:
\begin{equation*}
dm/dt = \alpha m - \beta mn, dn/dt = \delta mn - \gamma n
\end{equation*}
, where $m, n$ represent the population density of the prey and predator, respectively, and $\alpha, \beta, \delta, \gamma$ are the interaction parameters between the two species. 
The system state is defined as $x^e_t = (m^e_t, n^e_t) \in \mathbb{R}^2_+$ with initial conditions $(m_0, n_0)$ sampled from a uniform distribution $p(x_0) = Unif([1, 3]^2)$.
The environment $e$ is defined by dynamics parameters $\theta_e = (\alpha_e/\beta_e, \gamma_e/\delta_e) \in \Theta$, sampled uniformly from the set $\Theta$.
We simulate trajectories over a temporal grid with $\Delta t = 0.5$ and a horizon $T = 10$. 
At test time, environment labels are inferred from the prediction bias observed over an initial segment of the trajectory of length $\Delta t$.

\paragraph{Gray-Scott (GS)~\citep{pearson1993complex}} The model uses simple reaction-diffusion equations to effectively study complex pattern formation in chemical and biological systems, following underlying PDE dynamics:
\begin{equation*}
\begin{aligned}
& \partial m/ \partial t = D_m\Delta m - mn^2 + F(1-m), \\
& \partial n/ \partial t = D_n\Delta n - mn^2 - (F+k)n.    
\end{aligned}
\end{equation*}
, where $m, n$ represent the concentrations of two chemical components in the spatial domain $S$ with periodic boundary conditions, and $D_m, D_n$ are their constant diffusion coefficients, and $F, k$ are the reaction parameters that govern the spatio-temporal dynamic patterns. 
$S$ is a 2D space of dimension $32\times 32$ with spatial resolution of $\Delta s = 2$.
The system state $x^e_t = (m^e_t, n^e_t) \in \mathbb{R}^{2 \times 32^2}_+$. 
We define the initial conditions $(m_0, n_0)\sim p(x_0)$ by uniformly sampling three two-by-two squares,  which activate the reactions, from $S$. 
$(m_0, n_0) = (1-\epsilon, \epsilon)$ with $ \epsilon= 0.05$ inside the squares and $(m_0, n_0) = (0, 1)$ outside the squares.
The environment $e$ is defined by dynamics parameters $\theta = (F_e, k_e) \in \Theta$, sampled uniformly from the environment distribution $Q$ on $\Theta$.
We simulate trajectories on a temporal grid using a timestep of $\Delta t = 40$ over a horizon of $T = 400$. 
At test time, environment labels are inferred from an initial segment of length $\Delta t$.

\paragraph{Navier-Stokes (NS)~\citep{li2020fourier}} The Navier-Stokes PDE describes the motion of viscous fluid substances:

\begin{equation*}
\partial m/ \partial t = -n \nabla m + \nu \Delta m + \xi, \nabla v = 0
\end{equation*}
, where $n$ is the velocity field, $m = \nabla \times n$ is the vorticity, both $n,m$ lie in a spatial domain $S$ with periodic boundary conditions, $\nu$ is the viscosity (fixed as $1e^{-3}$) and $\xi$ is the constant forcing term in the domain $S$.
The system state is characterized by $x^e_t = m^e_t \in \mathbb{R}^{32^2}$, as initialized per~\citep{li2020fourier}.
The environment $e$ is determined by a uniformly sampled forcing term $\xi_e \in \Theta_\xi$.
We simulate trajectories across a temporal interval $\Delta t = 1$ over a horizon $T = 10$.
At test time, environment labels are inferred from an initial segment of length $2\Delta t$.

\label{appendix_environments}
The parameters for LV, GS and NS systems are respectively given in Table~\ref{tab:LV_params}, \ref{tab:GS_params} and \ref{tab:NS_params}.

\begin{table*}[htb]
\caption{Parameters of LV Systems}
\label{tab:LV_params}
\resizebox{\linewidth}{!}{
\begin{tabular}{c|ccccccccccc}
\toprule[1.5pt]
Params. & Train 1 & Train 2 & Train 3 & Train 4 & Train 5 & Train 6 & Train 7 & Train 8 & Train 9 & Adapt 1 & Adapt 2 \\ \hline
$\alpha$ & 0.5     & 0.5     & 0.5     & 0.5     & 0.5     & 0.5     & 0.5     & 0.5     & 0.5     & 0.7     & 0.6     \\
$\beta$  & 0.5     & 0.75    & 1       & 0.5     & 0.5     & 0.75    & 0.75    & 1       & 1       & 0.8     & 0.7     \\
$\gamma$ & 0.5     & 0.5     & 0.5     & 0.5     & 0.5     & 0.5     & 0.5     & 0.5     & 0.5     & 0.5     & 0.5     \\
$\delta$ & 0.5     & 0.5     & 0.5     & 0.75    & 1       & 0.75    & 1       & 0.75    & 1       & 0.5     & 0.5   \\
\bottomrule[1.5pt]
\end{tabular}}
\end{table*}

\begin{table*}[htb]
\caption{Parameters of GS Systems}
\label{tab:GS_params}
\centering
\resizebox{0.65\linewidth}{!}{
\begin{tabular}{c|ccccc}
\toprule[1.5pt]
Params. & Train 1 & Train 2 & Train 3 & Adapt 1 & Adapt 2 \\ \hline
$F$ & 0.037   & 0.03    & 0.039   & 0.033  & 0.036   \\
$k$ & 0.06    & 0.062   & 0.058   & 0.059   & 0.061  \\
\bottomrule[1.5pt]
\end{tabular}}
\end{table*}

\begin{table*}[htb]
\caption{Parameters of NS Systems}
\label{tab:NS_params}
\centering
\resizebox{0.65\linewidth}{!}{
\begin{tabular}{c|c}
\toprule[1.5pt]
& $\xi$           \\ \hline
Train 1 & $0.1 * (\sin(2\pi(X + Y)) + \cos(2\pi*(X + Y)))$   \\
Train 2 & $0.1 * (\sin(2\pi(X + Y)) + \cos(2\pi*(X + 2Y)))$  \\
Train 3 & $0.1 * (\sin(2\pi(X + Y)) + \cos(2\pi*(2X + Y)))$  \\
Train 4 & $0.1 * (\sin(2\pi(X + 2Y)) + \cos(2\pi*(2X + Y)))$ \\
Adapt 1 & $0.1 * (\sin(2\pi(2X + Y)) + \cos(2\pi*(X + 2Y)))$ \\
Adapt 2 & $0.1 * (\sin(2\pi(2X + Y)) + \cos(2\pi*(2X + Y)))$ \\
\bottomrule[1.5pt]
\end{tabular}}
\end{table*}

\section{Centroid-like NN Behavior}
We conducted an additional experiment measuring the MSE loss for each training sample across different neural networks (NNs). The results, shown in Table ~\ref{tab:mse_assignment}, yielded two key findings: first, each training sample has a uniquely best-fit NN where it attains minimal loss. Second, this optimal NN generalizes effectively to a broader cluster of trajectories—specifically, those originating from the same environment. This alignment between a "centroid-like" NN and environment-specific trajectory clusters strongly reinforces our methodological rationale.

\begin{table}[htbp]
\centering
\resizebox{\linewidth}{!}{
\begin{tabular}{ccccccccccc}
\toprule[1.5pt]
Traj. & NN1 & NN2 & NN3 & NN4 & NN5 & NN6 & NN7 & NN8 & NN9 & Assign. \\
\midrule
1  & 2.06e-01 & 1.18e-01 & 2.80e-01 & 7.76e-06 & 8.34e-02 & 1.59e-01 & 3.14e-01 & 2.50e-01 & 2.35e-01 & 3 \\
2  & 2.05e-01 & 1.21e-01 & 2.66e-01 & 9.35e-06 & 8.42e-02 & 1.53e-01 & 3.36e-01 & 2.40e-01 & 2.26e-01 & 3 \\
3  & 2.05e-01 & 1.21e-01 & 2.65e-01 & 1.03e-05 & 8.41e-02 & 1.52e-01 & 3.35e-01 & 2.39e-01 & 2.25e-01 & 3 \\
4  & 2.45e-01 & 1.51e-01 & 3.85e-01 & 1.98e-05 & 1.13e-01 & 1.89e-01 & 4.05e-01 & 3.10e-01 & 3.12e-01 & 3 \\
5  & 3.31e-02 & 1.65e-01 & 1.12e-01 & 8.32e-02 & 5.36e-06 & 5.42e-02 & 3.68e-01 & 7.26e-02 & 1.37e-01 & 4 \\
6  & 3.15e-02 & 1.87e-01 & 1.07e-01 & 8.46e-02 & 4.26e-06 & 5.57e-02 & 4.21e-01 & 6.80e-02 & 1.44e-01 & 4 \\
7  & 3.14e-02 & 1.86e-01 & 1.06e-01 & 8.46e-02 & 4.50e-06 & 5.52e-02 & 4.19e-01 & 6.74e-02 & 1.43e-01 & 4 \\
8  & 2.84e-02 & 2.68e-01 & 2.13e-01 & 1.14e-01 & 2.22e-05 & 9.57e-02 & 5.38e-01 & 1.18e-01 & 2.36e-01 & 4 \\
9  & 7.60e-06 & 2.32e-01 & 5.48e-02 & 2.07e-01 & 3.36e-02 & 4.44e-02 & 4.22e-01 & 1.91e-02 & 1.18e-01 & 0 \\
10 & 6.45e-06 & 2.64e-01 & 6.07e-02 & 2.05e-01 & 3.12e-02 & 5.17e-02 & 4.93e-01 & 2.10e-02 & 1.37e-01 & 0 \\
11 & 6.65e-06 & 2.63e-01 & 6.02e-02 & 2.05e-01 & 3.11e-02 & 5.14e-02 & 4.91e-01 & 2.07e-02 & 1.36e-01 & 0 \\
12 & 1.95e-05 & 3.66e-01 & 1.61e-01 & 2.45e-01 & 2.92e-02 & 9.83e-02 & 6.31e-01 & 7.05e-02 & 2.32e-01 & 0 \\
13 & 2.31e-01 & 1.35e-05 & 1.56e-01 & 1.16e-01 & 1.64e-01 & 8.52e-02 & 5.28e-02 & 1.84e-01 & 7.04e-02 & 1 \\
14 & 2.63e-01 & 1.39e-05 & 1.64e-01 & 1.22e-01 & 1.87e-01 & 9.48e-02 & 5.95e-02 & 2.02e-01 & 7.16e-02 & 1 \\
15 & 2.62e-01 & 1.41e-05 & 1.63e-01 & 1.22e-01 & 1.86e-01 & 9.44e-02 & 5.93e-02 & 2.01e-01 & 7.13e-02 & 1 \\
16 & 3.65e-01 & 1.89e-05 & 2.33e-01 & 1.50e-01 & 2.67e-01 & 1.19e-01 & 6.96e-02 & 2.56e-01 & 1.08e-01 & 1 \\
17 & 4.22e-01 & 5.35e-02 & 2.43e-01 & 3.15e-01 & 3.68e-01 & 1.95e-01 & 2.78e-05 & 3.17e-01 & 1.13e-01 & 6 \\
18 & 4.91e-01 & 5.91e-02 & 2.77e-01 & 3.35e-01 & 4.19e-01 & 2.27e-01 & 2.12e-05 & 3.66e-01 & 1.30e-01 & 6 \\
19 & 4.89e-01 & 5.89e-02 & 2.76e-01 & 3.34e-01 & 4.17e-01 & 2.26e-01 & 2.08e-05 & 3.64e-01 & 1.30e-01 & 6 \\
20 & 6.28e-01 & 6.90e-02 & 3.26e-01 & 4.04e-01 & 5.35e-01 & 2.56e-01 & 1.99e-05 & 4.18e-01 & 1.52e-01 & 6 \\
21 & 4.36e-02 & 8.60e-02 & 1.90e-02 & 1.58e-01 & 5.31e-02 & 7.09e-06 & 1.96e-01 & 1.98e-02 & 2.19e-02 & 5 \\
22 & 5.16e-02 & 9.53e-02 & 1.74e-02 & 1.54e-01 & 5.58e-02 & 8.08e-06 & 2.28e-01 & 2.09e-02 & 2.34e-02 & 5 \\
23 & 5.13e-02 & 9.49e-02 & 1.73e-02 & 1.53e-01 & 5.53e-02 & 7.86e-06 & 2.27e-01 & 2.08e-02 & 2.33e-02 & 5 \\
24 & 9.81e-02 & 1.19e-01 & 3.84e-02 & 1.89e-01 & 9.61e-02 & 1.45e-05 & 2.56e-01 & 2.72e-02 & 3.41e-02 & 5 \\
25 & 1.17e-01 & 7.14e-02 & 2.46e-02 & 2.35e-01 & 1.36e-01 & 2.11e-02 & 1.15e-01 & 5.47e-02 & 8.69e-06 & 8 \\
26 & 1.37e-01 & 7.21e-02 & 2.82e-02 & 2.26e-01 & 1.44e-01 & 2.36e-02 & 1.31e-01 & 6.42e-02 & 5.20e-06 & 8 \\
27 & 1.36e-01 & 7.18e-02 & 2.81e-02 & 2.26e-01 & 1.43e-01 & 2.35e-02 & 1.31e-01 & 6.39e-02 & 5.32e-06 & 8 \\
28 & 2.32e-01 & 1.08e-01 & 3.35e-02 & 3.12e-01 & 2.37e-01 & 3.47e-02 & 1.52e-01 & 7.46e-02 & 8.12e-06 & 8 \\
29 & 1.90e-02 & 1.84e-01 & 1.02e-02 & 2.49e-01 & 7.19e-02 & 2.00e-02 & 3.17e-01 & 2.50e-06 & 5.57e-02 & 7 \\
30 & 2.07e-02 & 2.03e-01 & 1.15e-02 & 2.41e-01 & 6.79e-02 & 2.11e-02 & 3.68e-01 & 2.91e-06 & 6.44e-02 & 7 \\
31 & 2.04e-02 & 2.02e-01 & 1.14e-02 & 2.40e-01 & 6.74e-02 & 2.10e-02 & 3.66e-01 & 3.05e-06 & 6.41e-02 & 7 \\
32 & 7.07e-02 & 2.55e-01 & 2.04e-02 & 3.09e-01 & 1.20e-01 & 2.74e-02 & 4.17e-01 & 8.74e-06 & 7.41e-02 & 7 \\
33 & 5.48e-02 & 1.56e-01 & 1.22e-06 & 2.80e-01 & 1.11e-01 & 1.89e-02 & 2.44e-01 & 1.00e-02 & 2.53e-02 & 2 \\
34 & 6.04e-02 & 1.65e-01 & 3.09e-06 & 2.67e-01 & 1.07e-01 & 1.76e-02 & 2.79e-01 & 1.14e-02 & 2.83e-02 & 2 \\
35 & 5.99e-02 & 1.64e-01 & 3.10e-06 & 2.66e-01 & 1.06e-01 & 1.75e-02 & 2.78e-01 & 1.14e-02 & 2.82e-02 & 2 \\
36 & 1.61e-01 & 2.32e-01 & 1.20e-05 & 3.85e-01 & 2.14e-01 & 3.90e-02 & 3.26e-01 & 2.06e-02 & 3.35e-02 & 2 \\
\bottomrule[1.5pt]
\end{tabular}
}
\caption{Training MSE for Trajectories by each Neural Network and Label Assignment}
\label{tab:mse_assignment}
\end{table}






\clearpage
\section{Environment Assignment Convergence on GS}
\begin{figure}[h!]
\centering
\includegraphics[width = \linewidth, keepaspectratio]{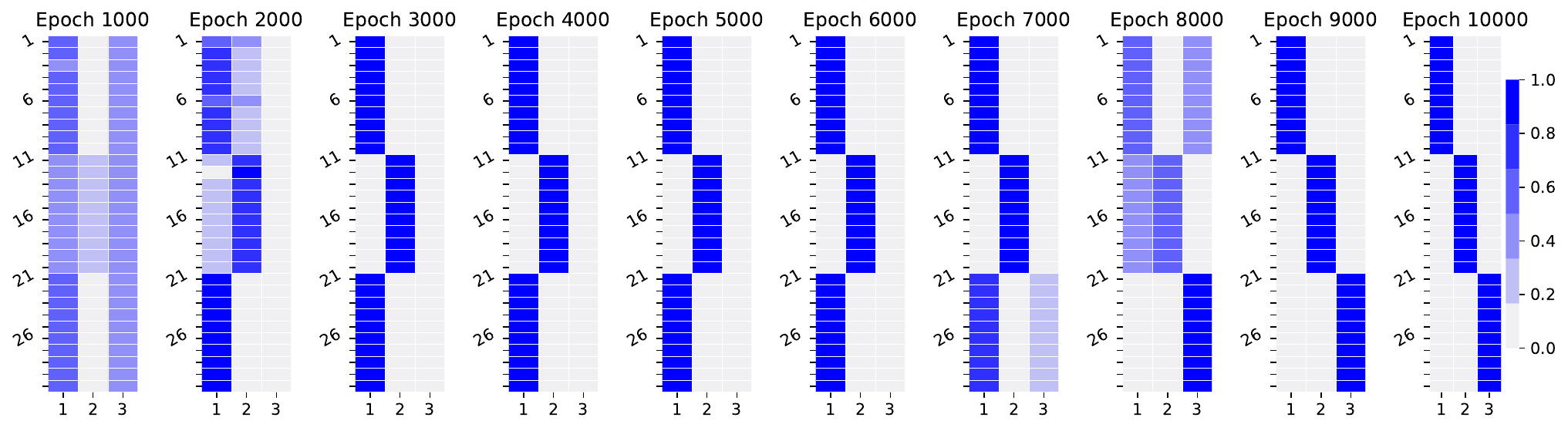}
\caption{Environment Assignment Probability over Time with LEADS as Base Model on GS. Despite initial inaccuracies due to complex dynamics, the assignment ultimately converges to the correct label.
}
\label{fig_assignment2}
\end{figure}

Figure~\ref{fig_assignment2} presents the temporal evolution of environment assignment probabilities using the LEADS base model on GS.
\label{sec:fig_conv_gs}

\section{Plots on Learned Dynamics}
\label{appendix_plot_dynamic}
\begin{figure*}[htb]
\centering
\includegraphics[width=0.5\linewidth]{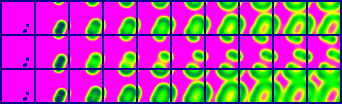}

\vspace{5pt}
\includegraphics[width=0.5\linewidth]{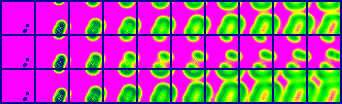}

\vspace{5pt}
\includegraphics[width=0.5\linewidth]{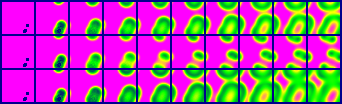}
\caption{Comparison of final GS states predicted by \model~(bottom) and Oracle (middle) against the ground truth (top) with base model LEADS.}
\label{fig_dyn_gs_leads}
\end{figure*}

\begin{figure*}[htb]
\centering
\includegraphics[width=0.5\linewidth]{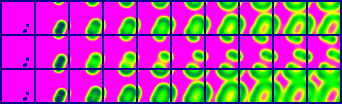}

\vspace{5pt}
\includegraphics[width=0.5\linewidth]{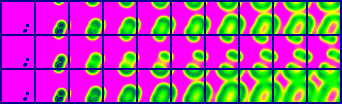}

\vspace{5pt}
\includegraphics[width=0.5\linewidth]{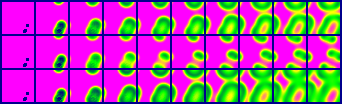}
\caption{Comparison of final GS states predicted by \model~(bottom) and Oracle (middle) against the ground truth (top) with base model CoDA-$l_1$.}
\label{fig_dyn_gs_coda}
\end{figure*}

\begin{figure*}[htb]
\centering
\includegraphics[width=0.5\linewidth]{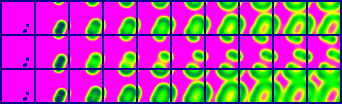}

\vspace{5pt}
\includegraphics[width=0.5\linewidth]{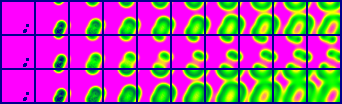}

\vspace{5pt}
\includegraphics[width=0.5\linewidth]{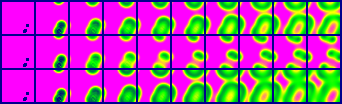}
\caption{Comparison of final GS states predicted by \model~(bottom) and Oracle (middle) against the ground truth (top) with base model CoDA-$l_2$.}
\label{fig_dyn_gs_coda2}
\end{figure*}

\begin{figure*}[htb]
\centering
\includegraphics[width=0.5\linewidth]{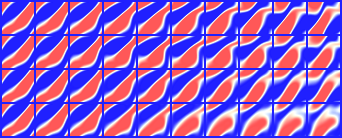}

\vspace{5pt}
\includegraphics[width=0.5\linewidth]{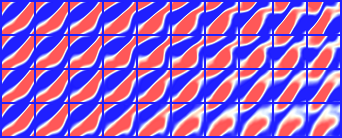}

\vspace{5pt}
\includegraphics[width=0.5\linewidth]{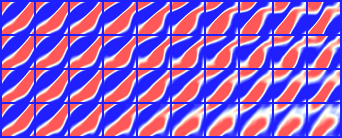}
\caption{Comparison of final NS states predicted by \model~(bottom) and Oracle (middle) against the ground truth (top) with base model LEADS.}
\label{fig_dyn_ns_leads}
\end{figure*}

\begin{figure*}[htb]
\centering
\includegraphics[width=0.5\linewidth]{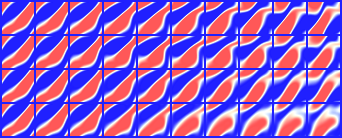}

\vspace{5pt}
\includegraphics[width=0.5\linewidth]{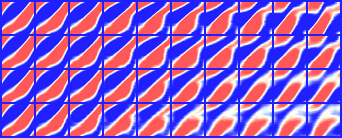}

\vspace{5pt}
\includegraphics[width=0.5\linewidth]{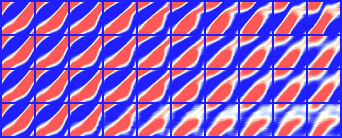}
\caption{Comparison of final NS states predicted by \model~(bottom) and Oracle (middle) against the ground truth (top) with base model CoDA-$l_1$.}
\label{fig_dyn_ns_coda}
\end{figure*}

\begin{figure*}[htb]
\centering
\includegraphics[width=0.5\linewidth]{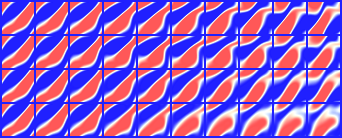}

\vspace{5pt}
\includegraphics[width=0.5\linewidth]{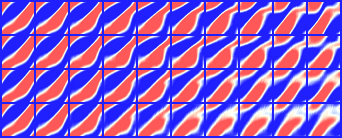}

\vspace{5pt}
\includegraphics[width=0.5\linewidth]{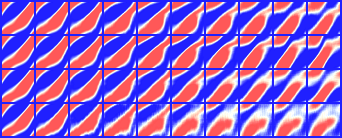}
\caption{Comparison of final NS states predicted by \model~(bottom) and Oracle (middle) against the ground truth (top) with base model CoDA-$l_2$.}
\label{fig_dyn_ns_coda2}
\end{figure*}

\begin{figure*}[htb]
\centering
\includegraphics[width=0.6\linewidth]{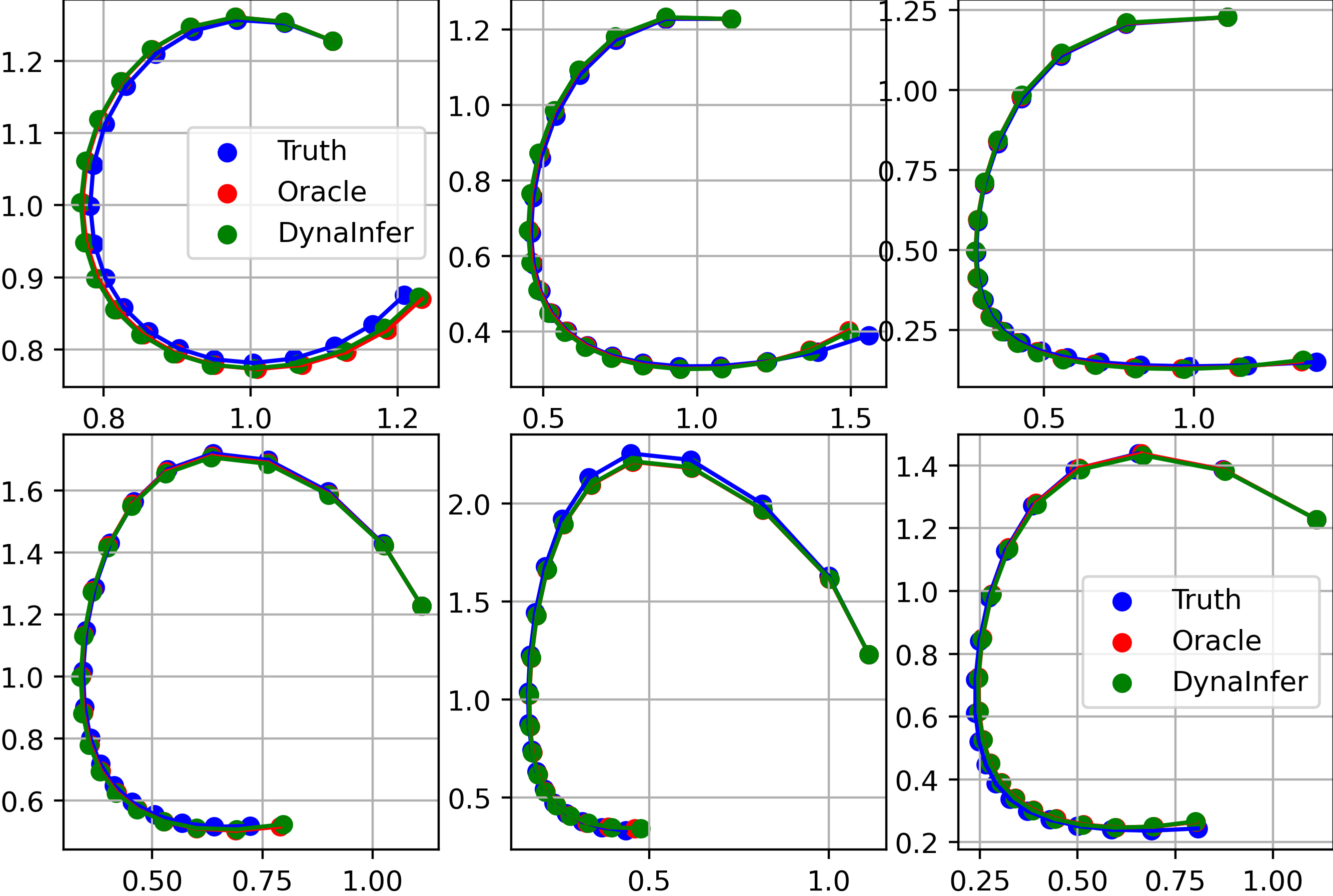}
\hrule
\vspace{10pt}
\includegraphics[width=0.6\linewidth]{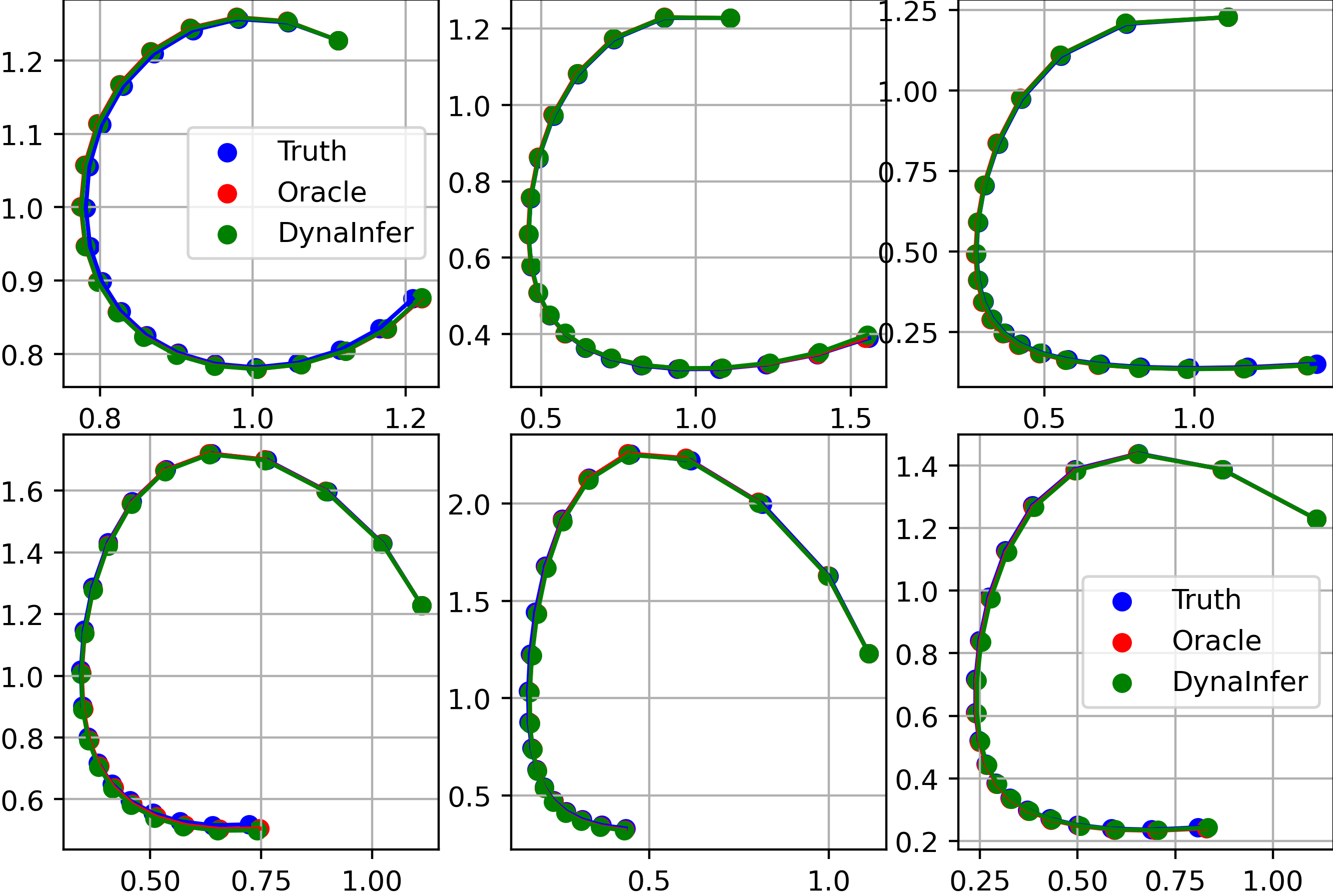}
\hrule
\vspace{10pt}
\includegraphics[width=0.6\linewidth]{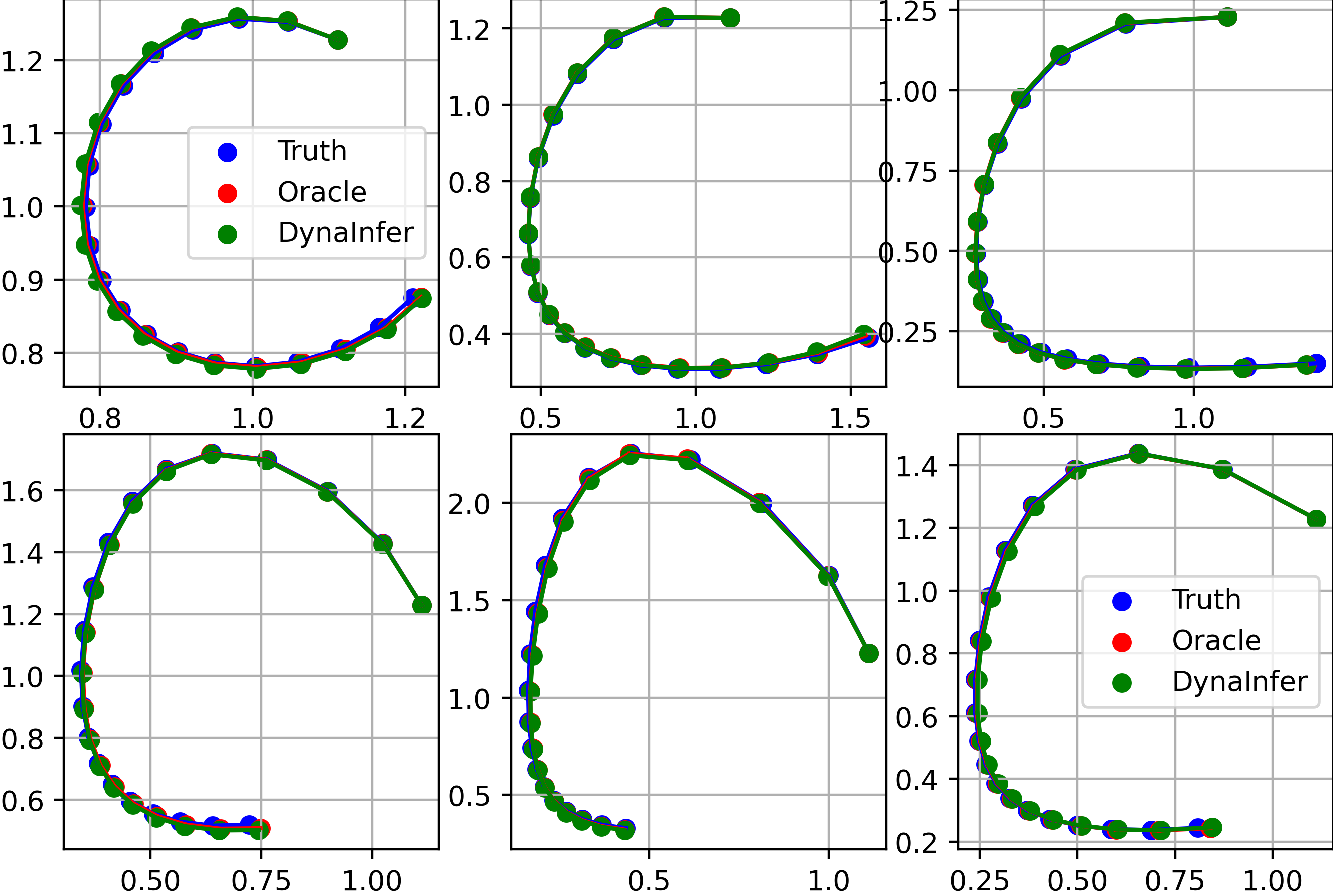}
\caption{Comparison of predicted LV trajectories from 6 environments by \model~(green) and Oracle (red) against the ground truth (blue) with base models LEADS (top), CoDA-$l_1$ (middle), and CoDA-$l_2$ (bottom).}
\label{fig_dyn_lv}
\end{figure*}

We present the recovered test trajectories produced by the learned neural network. Figures~\ref{fig_dyn_gs_leads}, \ref{fig_dyn_gs_coda}, and \ref{fig_dyn_gs_coda2} illustrate GS; Figures~\ref{fig_dyn_ns_leads}, \ref{fig_dyn_ns_coda}, and \ref{fig_dyn_ns_coda2} depict NS; and Figure~\ref{fig_dyn_lv} shows LV. A close examination reveals that the trajectories predicted by \model~closely align with both the Oracle and the ground truth for the selected systems.

\end{document}